\documentclass[sn-mathphys, pdflatex]{sn-jnl}

\theoremstyle{thmstyleone}%
\newtheorem{theorem}{Theorem}
\newtheorem{proposition}[theorem]{Proposition}%

\theoremstyle{thmstyletwo}%
\newtheorem{remark}{Remark}%

\theoremstyle{thmstylethree}%

\usepackage{bm, bbm}

\newcommand{\bx}{\ensuremath{\boldsymbol{x}}}

\newcommand{\comment}[1]{}

\newcommand{\veps}{\varepsilon}
\newcommand{\bbN}{\mathbb{N}}

\newcommand{\rev}[1]{{#1}}

\definecolor{grey}{rgb}{0.75,0.75,0.75}
\definecolor{orange}{rgb}{1.0,0.5,0.5}
\definecolor{brown}{rgb}{0.5,0.25,0.0}
\definecolor{pink}{rgb}{1.0, 0.5,0.5}
\newtheorem{lemma}{Lemma}

\newtheorem{Definition}{Definition}
\numberwithin{equation}{section}

\begin{document}

 \title[ChebNet: RePU Deep Neural Networks via Chebyshev
 Approximation]{ChebNet: Efficient and Stable Constructions of Deep Neural
 Networks
 with Rectified Power Units via Chebyshev Approximation}

   \author[1]{\fnm{Shanshan} \sur{Tang}}\email{tangss@sdc.icbc.com.cn}
   \equalcont{These authors contributed equally to this work.
       Part of the work of the two authors are done during their
   Ph.D. study at LSEC, Academy of Mathematics and Systems Science, CAS.}

   \author[2]{\fnm{Bo} \sur{Li}}\email{libo296@hisilicon.com}
   \equalcont{These authors contributed equally to this work.
       Part of the work of the two authors are done during their
   Ph.D. study at LSEC, Academy of Mathematics and Systems Science, CAS.}

   \author*[3,4]{\fnm{Haijun} \sur{Yu}}\email{hyu@lsec.cc.ac.cn}

   	\affil[1]{Software Development Center, Industrial and Commercial Bank of
       China, No. 16 Building of ZhongGuanCun Software Park, Haidian District,
       Beijing
       100193,
       China}
   \affil[2]{Hisilicon Semiconductor and Component Business Dept.(2012 Labs),
       Huawei Technologies Co., Ltd, Bai Ruida Apartment, Bantian Street,
       Longgang
       District, Shenzhen 518129, China}

   \affil*[3]{NCMIS \& LSEC, \orgdiv{Institute of Computational Mathematics and
   Scientific/Engineering Computing},
   	\orgname{Academy of Mathematics and Systems Science}, \city{Beijing}
   	\postcode{100190}, \country{China}}
   \affil*[4]{\orgdiv{School of Mathematical Sciences}, \orgname{University of
   Chinese Academy of Sciences}, \city{Beijing} \postcode{100049},
   \country{China}}

\abstract{
     In a previous study [B. Li, S. Tang and H. Yu, Commun. Comput. Phy.
     27(2):379-411, 2020], it is shown that deep neural networks built with
     rectified power units (RePU) as activation functions can give better
     approximation for sufficient smooth functions than those built with
     rectified linear units, by converting
     polynomial approximations using power series into deep neural
     networks with optimal complexity and no approximation
     error. However, in practice, power series approximations are not easy to
     obtain due to
     the associated stability issue.
     In this paper, we propose a new and more stable way to construct
     RePU deep neural networks based on Chebyshev polynomial
     approximations.  By using a hierarchical structure of Chebyshev
     polynomial approximation in frequency domain, we obtain efficient
     and stable deep neural network construction, which we call ChebNet.
     The approximation of smooth functions by ChebNets is no worse than
     the approximation by deep RePU nets using power series. On the same time,
     ChebNets are much more stable. Numerical results show that
     the constructed ChebNets can be further fine-tuned to obtain much
     better results than those obtained by tuning deep RePU nets
     constructed by power series approach. 
 	As spectral accuracy is hard to obtain by direct training of deep neural networks, ChebNets provide a practical way to obtain spectral accuracy, it is expected to be useful 
 in real applications that require efficient approximations of smooth functions.}

  \pacs[MSC Classification]{65D15, 68T07, 41A10}

\keywords{Deep neural networks, Rectified power units, Chebyshev polynomial,
  ChebNet, Stability, Constructive approximation}

\maketitle
\section{Introduction}

Deep neural networks (DNNs), which compose multi-layers of affine
transforms and nonlinear activations, is getting more and more popular
as a universal modeling tool since the seminal works
\citep{hinton_fast_2006} and \citep{bengio_greedy_2007}.  DNNs have
greatly boosted the developments in different areas including image
classification, speech recognition, computational chemistry, numerical
solutions of high-dimensional partial differential equations and other
scientific problems, see e.g. \cite{hinton_deep_2012,
  krizhevsky_imagenet_2012, lecun_deep_2015,he2016deep,
  zhang_deep_2018,e_deep_2018,
  han_solving_2018,kutyniok_theoretical_2019} and the references
therein.

One of the basic facts behind the success of DNNs is that DNNs are
universal approximators. It is well-known that neural networks with only
one hidden layer can approximate any $C^0$ or $L^1$ functions with any
given error tolerance
\citep{cybenko_approximation_1989,hornik_multilayer_1989}.  In fact, for
neural networks with only one-hidden layer of non-polynomial $C^\infty$
activation functions, Mhaskar \cite{mhaskar_neural_1996} proved that the upper
error bound of approximating multi-dimensional functions is of spectral
type: Error rate $\veps\sim n^{-k/d}$ can be obtained theoretically for
approximating functions in Sobolev space $W^k([-1, 1]^d)$. Here $d$ is
the number of dimensions, $n$ is the number of hidden nodes in the
neural network.  Due to the success of DNNs, people believe that deep
neural networks have broader scopes of representation than shallow
ones. Recently, several works have demonstrated or proved this in
different settings, see
e.g. \cite{telgarsky_representation_2015,eldan_power_2016,poggio_why_2017}.
One of the commonly used activation functions with DNNs is the so-called
rectified linear unit (ReLU), 
which is defined as $\sigma(x)=\max(0,x)$.
Telgarsky \cite{telgarsky_representation_2015} gave a simple and elegant
construction showing that for any $k$, there exist $k$-layer,
$\mathcal{O}(1)$ wide ReLU networks on one-dimensional data, which can
express a sawtooth function on $[0,1]$ with $\mathcal{O}(2^k)$
oscillations. Moreover, such a rapidly oscillating function cannot be
approximated by poly$(k)$-wide ReLU networks with $o(k/\log(k))$ depth.
Following this approach, several other works proved that deep ReLU
networks have better approximation power than shallow ReLU networks, see
e.g.  \cite{liang_why_2016}, \cite{telgarsky_benefits_2016a},
\cite{yarotsky_error_2017}, \cite{petersen_optimal_2018}.  In
particular, for $C^\beta$-differentiable $d$-dimensional functions,
\cite{yarotsky_error_2017} proved that the number of parameters needed
to achieve an error tolerance of $\varepsilon$ is
$\mathcal{O}(\varepsilon^{-\frac{d}{\beta}}\log\frac{1}{\varepsilon})$.
\cite{petersen_optimal_2018} proved that for a class of $d$-dimensional
piecewise $C^\beta$ continuous functions with the discontinuous
interfaces being $C^\beta$ continuous also, one can construct a ReLU
neural network with $\mathcal{O}((1+\frac{\beta}{d})\log_2 (2+\beta) )$
layers, $\mathcal{O}(\varepsilon^{-\frac{2(d-1)}{\beta}})$ nonzero
weights to achieve $\varepsilon$-approximation. The complexity bound is
sharp. The spectral convergence of using deep ReLU network approximating
analytic functions was proved by \cite{e_exponential_2018} and
\cite{opschoor_exponential_2019}.  The significance of the above
mentioned works is that by using a very simple rectified nonlinearity,
DNNs can obtain high order approximation property. Shallow networks do
not hold such a good property.

A key fact used in the error estimates of deep ReLU networks is that
$x^2, xy$ can be approximated by a ReLU network with
$\mathcal{O}(\log\frac1\veps)$ layers, which introduces a
$\log\frac1\veps$ factor or a big constant related to the smoothness of
functions to be approximated.  To remove the approximation error and the
extra $\log\frac1\veps$ factor in the size of neural networks,
We proposed in \cite{li_better_2019} to use rectified power units (RePU) to
construct exact neural network representations of polynomials with
optimal size. The RePU function is defined as
\begin{equation}
\label{eq:RePU}
\sigma_s(x) = \begin{cases}
x^s, & x \ge 0, \\
0,   & x<0,
\end{cases}
\end{equation}
where $s$ is a non-negative integer. {For $s=2, 3$, the function
  $\sigma_2, \sigma_3$ are called ReQU and ReCU, respectively.}  Note
that, the RePU functions have been used by
Mhaskar \cite{mhaskar_approximation_1993} as activation functions to construct
neural networks based on spline approximation. Using Mhaskar's
construction, a polynomial with degree $n$ will be converted into a RePU
network of size $\mathcal{O}(n\log n)$ with depth $\mathcal{O}(\log
n)$. By using a different approach, we gave optimal
and stable constructions to convert a polynomial of degree $n$ into a
$\sigma_2$ network of size $\mathcal{O}(n)$ with depth
$\mathcal{O}(\log n)$ \cite{li_better_2019}. Similar constructions are extended
to
$\sigma_s(s\ge 2)$ DNNs in \cite{li_PowerNet_2019}. Combining with
classical polynomial approximation theory, the constructed deep RePU
networks can approximate smooth functions with spectral
accuracy. {For example, the following theorem for approximating
  functions in Jacobi-weighted Sobolev space is proved in
  \cite{li_better_2019}:

\begin{theorem}\label{thm:MdSobolev}
  For any $u\!\in\! {B^{m}_{\bm{\alpha},\bm{\beta}}}(I^d)$, with
  $\vert u \vert_{B^m_{\bm{\alpha},\bm{\beta}}(I^d)} \le 1$, $
  \bm{\alpha,\beta}\!\in\!(-1,\infty)^{d}$, and any
  $\varepsilon\in(0,1)$ there exists a $\sigma_2$ neural network
  $\Phi_\varepsilon^u$ having
  $\mathcal{O}\left(\frac{d}{m} \log_2 \frac{1}{\varepsilon} + d\right)$
  layers with no more than $\mathcal{O}\left(\varepsilon^{-d/m}\right)$
  nodes and nonzero weights, that approximates $u$ with
  ${L^2_{\omega^{\bm{\alpha},\bm{\beta}}}(I^d)}$-error less than
  $\varepsilon$, i.e.
	\begin{equation}
	\| R_{\sigma_2} (\Phi_\varepsilon^u) - u \|_{L^2_{\omega^{\bm{\alpha},\bm{\beta}}}(I^d)}
	\le \varepsilon.
	\end{equation}
\end{theorem}

Here $I^d:=[-1,1]^d$, and the
Jacobi-weighted Sobolev
space $B_{\bm{\alpha},\bm{\beta}}^m(I^d)$ for $ \ m \in {\bbN}_0$,  is defined
as
\begin{equation}
\label{eq:wtSobolevSpaceMd}
B_{\bm{\alpha},\bm{\beta}}^m(I^d) := \left\{\,u\in L^2(I^d) \mid
    \partial^{\bm{k}}_{\bx} u 
    \in L^2_{\omega^{\bm{\alpha}+\bm{k}, \bm{\beta}+\bm{k}} }(I^d),\,
    \bm{k}\in {\bbN}_0^d,\,
    \vert \bm{k}\vert_{1}\le m\, \right\},
\end{equation}
where $\omega^{\bm{\alpha},\bm{\beta}}(\bx) :=
\omega^{\alpha_1,
	\beta_1}(x_1)\cdots\omega^{\alpha_d, \beta_d}(x_d)$ is the multi-dimensional Jacobi weight.
}

Moreover, RePU networks have other good properties. E.g. functions
  represented by $\sigma_s (s\ge2)$ networks are smooth functions, thus
  fit the situations where derivatives of network are involved in the
  loss function, e.g. DeepRitz\cite{e_deep_2018},
  PINN\cite{raissi_physicsinformed_2019}, FNM\cite{xu_finite_2020a},
  OnsagerNet\cite{yu_onsagernet_2021}, MIM\cite{lyu_mim_2022}; comparing
  to ReLU networks, the sizes and depths of ReQU
  networks needed to approximate smooth functions with certain accuracy are
  smaller, so they suit the situation where a lot of evaluations are
  needed.

However, there is one drawback in the RePU networks constructed in
\citep{li_better_2019} and \citep{li_PowerNet_2019}, where polynomial
approximations based on power series are used. There are two ways to
calculate the power series to represent a function. The first one is to
use Taylor expansion to approximate the power series, which might
diverge if the radius of convergence of the Taylor series is not large
enough, and the high order derivatives used in Taylor series are not
easy to calculate. The second approach first approximates the function
using some orthogonal polynomial projection or interpolation, then
converts the resulting polynomial approximation into power series. But
in this approach, the condition number associated to the transforms from
orthogonal polynomial bases to monomial bases are known grows exponentially
fast (see, e.g. \citep{gautschi_optimally_2011}). { The transform corresponds to
  solve a linear problem $A z = c$, where $c$ is the coefficients of
  orthogonal expansion, $z$ is the coefficients of power series. It is
  well-known the the error in $c$ and error in $z$ satisfy the following
  estimate
\begin{equation}
 \label{eq:condnum_err}
 \frac{\| \delta z\|}{ \| z \|} \le \kappa(A) \frac{\| \delta c\|}{\|c\|}.
 \end{equation}
 So a small error in $c$ could be enlarged in $z$ due to the big
 condition number.  Even though for some special problems, relative
 errors independent of the condition number could be obtained
 \citep{higham_error_1987}, the matrix $A$ with big condition number
 will change the relative scaling of the coefficients. When there are
 big coefficients with opposite signs, the cancellation will reduce the
 number of significant digits. In such a case, the accuracy given in Theorem
 \ref{thm:MdSobolev} will deteriorate for numerical implementations using any
 float-point
 number system. }

In this paper, we propose a new construction of deep RePU networks to
remove the drawback mentioned above. {W}e accomplish this goal by
constructing deep RePU networks based on Chebyshev polynomial
approximation directly.  Chebyshev method is one of the most popular
spectral methods for function approximation and solving partial
differential equations, see
e.g. \cite{boyd_chebyshev_2001}, \cite{platte_chebfun_2010}.  It is
known that Chebyshev approximation can be efficiently calculated by
using fast Fourier transforms. By using a hierarchical structure of
Chebyshev polynomial approximation in frequency domain, we develop
methods in this paper to convert Chebyshev polynomial approximations
into deep RePU networks with optimal sizes, we call the resulting neural
networks ChebNets.\footnote[1]{There is another work involves using Chebyshev
  polynomial in graph convolutional
  networks\citep{defferrard_convolutional_2016}, which called
  ChebyNet. The two networks are very different, should not be confused.}
  Correspondingly, the
RePU networks constructed using methods proposed in
\citep{li_better_2019} and \citep{li_PowerNet_2019} will be referred to
as PowerNets in this paper.  ChebNets have all the good theoretical
approximation
properties that PowerNets have: they have faster convergence in
approximating sufficient smooth functions than deep ReLU networks; they
fit in situations where derivatives involved in the loss function, for
which deep ReLU networks are hard to use.
In addition, ChebNets provide
several distinct benefits:
    \begin{enumerate}
        \item ChebNets are numerically more stable, which can provide
          better results by increasing the degree of underlying
          polynomial approximation to represent functions with finite
          order of smoothness. In contrast, the stability issue makes
          the performance of PowerNets deteriorate quickly as the
          number of terms in power series increases.

        \item Even if the stability issue is not a serious problem for
          some particular functions, the ChebNet approach is still
          better than the PowerNet approach in terms of computational
          efficiency, since Chebyshev expansion coefficients can be
          stably and efficiently calculated by fast Fourier transform
          in $\mathcal{O}(n\log n)$ operations, while numerically
          calculating the coefficients of power expansion usually
          needs $\mathcal{O}(n^2)$ operations.

        \item ChebNets can be further fine-tuned to obtain much better
          results than those obtained by tuning PowerNets, which will be
          demonstrated by the numerical results given in Section 4.
    \end{enumerate}
With above mentioned good properties: high order accuracy, stable
structure, and optimal complexity in both network construction and
forward propagation, we expect that ChebNets will be a good choice to
be embedded into big neural networks in some real applications to get
better performance.

The remaining part of this paper is organized as follows. We first
present the optimal deep RePU network constructions (ChebNets) to exactly
represent Chebyshev polynomial approximations in Section 2. We
then discuss the approximations of smooth functions with ChebNets in
Section 3. In Section 4, we present some numerical experiments and
explain why ChebNets are more stable than PowerNets. We finish the paper
by a short summary in Section 5.

\section{Construction of ChebNets}

In this section, we construct deep RePU networks to represent
Chebyshev expansions exactly.

We first introduce notations. Let $\mathbb{N}$ denote all positive
integers, $\mathbb{N}_0 := \{0\}\cup \mathbb{N}$, and
$d,L\in \mathbb{N}$, then a neural network $\Phi$ with $d$ dimensional
input, $L$ layers can be denoted with a matrix-vector sequence
\begin{equation}
\Phi=\big( (A_1,b_1),\ldots,(A_L,b_L) \big),
\end{equation}
where $A_k$ are $N_k\times N_{k-1}$ matrices, $b_k\in\mathbb{R}^{N_k}$, {$N_{0}=d$,} $N_k\in \mathbb{N}$, $k=1,\cdots, L$. Let $\rho: \mathbb{R}\longrightarrow \mathbb{R}$ be
a nonlinear function that served as activation units, we define
\begin{equation}
 R_{\rho}(\Phi): \mathbb{R}^d \to \mathbb{R}^{N_L},\quad
 R_{\rho}(\Phi)(\bm{x})=\bm{x}_L,
\end{equation}
where $R_{\rho}(\Phi)(\bm{x})$ is defined as
\begin{align}
& \bm{x}_0:=\bm{x},\\
& \bm{x}_k=\rho(A_k\bm{x}_{k-1}+b_k),\quad k=1,\ldots,L-1,\\
& \bm{x}_L=A_L\bm{x}_{L-1}+b_L,
\end{align}
and
\begin{equation}
\rho(\bm{y})=(\rho(y^1),\ldots,\rho(y^m)), \quad \forall\,
\bm{y}=(y^1,\ldots,y^m)\in \mathbb{R}^m.
\end{equation}
We define the depth of the network $\Phi$ as the number of hidden
layers, which equals $L-1$. The number of nodes in each hidden layer is
$N_k(k=1,\ldots,L-1)$, and the summation of numbers of non-zero weights in
$A_k,b_k(k=1,\ldots,L)$ gives the total number of nonzero weights. In
this paper, we use these three quantities (number of hidden layers,
number of activation functions, and total number of nonzero weights) to
measure the complexity of a neural network.

Next, we list some basic facts about Chebyshev polynomials and RePU ($s=2$) realization.
\begin{lemma}\label{lem:1D}
For Chebyshev polynomials of the first kind $T_n(x)$, $x\in [-1,1]$, the following relations hold:
\begin{align}
& T_n(x) = \cos(n\arccos(x)), \qquad \qquad {T_{mn}(x) = T_m(T_n(x))}, \label{eq:powerrel} \\
& T_0(x)=1,\qquad T_1(x)=x,\qquad\qquad  T_{n+1}(x)=2xT_n(x)-T_{n-1}(x),
\quad n\ge 1,
\label{eq:cheb_3term}\\
& T_{m+n}(x)=2T_m(x)T_n(x)-T_{{\vert m-n \vert}}(x),\quad
  T_{2n}(x)=2T_n^2(x)-1,
  \label{eq:cheb_dbl_rel}\\
& \int_{-1}^{1} T_n(x)T_m(x)\omega(x)dx=\frac{h_n\pi}{2}\delta_{nm}, \label{eq:cheb_orth}
\end{align}
where $m,n\in \mathbb{N}_0$, $\omega(x)=(1-x^2)^{-\frac{1}{2}}$, $h_0=2$, $h_n=1$ for $n \ge 1$. $\delta_{nm}$ is the Kronecker delta, its value is $1$ if $m=n$ and $0$ otherwise.
\end{lemma}

\begin{lemma}[Lemma 1 in \cite{li_better_2019}]\label{lem:2}
For any $x, y\in \mathbb{R}$, the following identities hold:
 \begin{align}
 x  & =\beta_1^T\sigma_2(\omega_1x+\gamma_1), \\
 x^2& =\beta_2^T\sigma_2(\omega_2 x),\\
 xy & =\beta_1^T\sigma_2(\omega_1x+\gamma_1y), \label{eq:ReQU_idx}
 \end{align}
 where
$\beta_1=\frac{1}{4}[1,1,-1,-1]^T,\ \beta_2=[1,1]^T$,
$\omega_1=[1,-1,1,-1]^T$,
$\omega_2=[1,-1]^T$,
$\gamma_1=[1,-1,-1,1]^T$.

\end{lemma}

\begin{remark}
The Chebyshev polynomial $T_0(x) = 1$ can be realized by a
one layer neural network $\Phi\!=\!((0,1))$ without activation functions.
Based on Lemma \ref{lem:2}, $T_1(x)$ and $T_2(x)$ can be realized using one layer of $\sigma_2$ nodes as
\begin{align}
 T_1(x) &=\beta_1^T\sigma_2(\omega_1x+\gamma_1)\label{eq:T1}\\
 T_2(x) &=2\beta_2^T\sigma_2(\omega_2x) - 1.  \label{eq:T2}
 \end{align}
\end{remark}

\begin{theorem}
  \label{thm:RePUTn}
  For  $m\in \mathbb{N}$, the Chebyshev polynomial $T_{2^m}(x)$ of degree $2^m$
  defined on
  $\mathbb{R}$ can be represented exactly by a deep $\sigma_2$ network.
   The number of hidden layers, number of hidden nodes and number of nonzero
   weights required are at most $m$, $2m$ and $6m$, respectively.
  On the other sides, $T_{2^m}(x)$ can not be represented exactly
   by any $\sigma_2$ network with less than $m$ hidden layers.
\end{theorem}
\begin{proof}
    For the first part of the theorem, we give a constructive proof.
    By using property $T_{mn}(x)=T_m (T_n(x))$, we have
    $T_{2^m}(x)=(T_2\circ)^m (x)$. $T_2(x)$ can be represented exactly by a
    $\sigma_2$ neural network according to \eqref{eq:T2}. By stacking up $m$
    copies of $T_2$ networks, we obtain a deep $\sigma_2$ network to exactly
    represent $T_{2^m}$. By construction, the number of hidden layers is $m$,
    the number of hidden nodes is $2m$, and the number of nonzero coefficients is less than
   $6(m-1)+5$. The first part is proved.
   The second part is obvious, since $\sigma_2$ network with one hidden layer
   represent a piecewise polynomials of degree up to 2, applying piecewise
   polynomials
   of degree up to $m$ on a piecewise polynomial of degree up to $n$ leads to a
   piecewise
   polynomials of degree no more than $mn$. So, $\sigma_2$ network with hidden
  layers less than $m$  can not exactly represent any polynomial of degree $2^m$.
\end{proof}

\begin{remark}
  Due to the fact that Chebyshev polynomials are orthogonal polynomials on
  the interval $[-1, 1]$ with respect to the weight $\frac{1}{\sqrt{1-x^2}}$, $T_n(x)$
  has $n$ distinct zero points in $[-1,1]$, i.e. it has $n$ monotone regions
  or $n-1$ oscillations. Thus, a deep $\sigma_2$ neural networks with $m$ hidden layers
  and $\mathcal{O}(m)$ hidden nodes can produce
  functions with $\mathcal{O}(2^m)$ oscillations. Similar results can be
  extended to $\sigma_s$ neural networks. On the other sides, since
  $\sigma_s$ is a monotone function, a shallow $\sigma_s$ network with $m$ hidden
  nodes defined on $\mathbb{R}$ produce at most $\mathcal{O}(m)$ oscillations.
\end{remark}

\subsection{ChebNets for univariate Chebyshev polynomial expansions}

\begin{theorem}\label{thm:1D}
For $n\ge 1$, assume $p(x)=\sum_{j=0}^{n}c_jT_{j}(x)$, with $c_n\neq 0$, $x\in \mathbb{R}$, then there exists a $\sigma_2$ neural network with at most $\lfloor\log_2 n\rfloor$+1 hidden layers to represent $p(x)$ exactly. The number of neurons and total non-zero weights are both $\mathcal{O}(n)$.
\end{theorem}
\begin{proof}
    We split the proof into several steps.
\begin{itemize}
\item [(1)] For $n=1$, by Lemma \ref{lem:2}, we have
\begin{equation}
p(x)=c_0+c_1T_1(x)=c_0+c_1\beta_1^T\sigma_2(\omega_1 x+\gamma_1), \label{eq:p1}
\end{equation}
which shows  $p(x)$ can be represented exactly by a $\sigma_2$ network with one hidden layer. Written in the form
$ p(x)=A_2 \sigma_2(A_1x+b_1) + b_2, $
we have
\[
A_1=\omega_1,\quad b_1=\gamma_1, \qquad
A_2=c_1\beta_1^T,\quad b_2=c_0.
\]

\item [(2)] For $n=2$, by Eq. \eqref{eq:T1} and \eqref{eq:T2}, we have
\begin{align*}
p(x) &=c_0\!+\!c_1T_1(x)+\!c_2T_2(x) \\
&= ( c_0\! -\! c_2 ) + c_1 \beta_1^T\sigma_2(\omega_1x+\gamma_1)
+ 2c_2 \beta_2^T\sigma_2(\omega_2x).
\end{align*}
This is a $\sigma_2$ network with one hidden layer of the form
$p(x)=A_2 \sigma_2(A_1x+b_1) + b_2$ with
\[
A_1=
\begin{bmatrix}
\omega_1\\
\omega_2
\end{bmatrix},\quad
b_1=
\begin{bmatrix}
\gamma_1\\
\bm{0}
\end{bmatrix}, \quad
A_2=[c_1\beta_1^T\ 2c_2\beta_2^T], \quad b_2= c_0-c_2.
\]

\item [(2)] For $n=3$, by Lemma \ref{lem:1D}, we have
 \begin{align*}
 p(x)& =c_0+c_1T_1(x) +c_2T_2(x) +c_3T_3(x)\\
     & =c_0+(c_1-c_3)T_1(x) + T_2(x)(c_2+2c_3T_1(x)),
 \end{align*}
showing that a network with two hidden layers can represent $p(x)$ exactly. Details are shown in the following.

\quad Immediate variables of the first hidden layer is
\begin{align*}
& \xi_1^{(1)}=c_0+(c_1-c_3)T_1(x)=c_0+(c_1-c_3)\beta_1^T\sigma_2(\omega_1x+\gamma_1),\\
&
\xi_1^{(2)}=c_2+2c_3T_1(x)=c_2+2c_3\beta_1^T\sigma_2(\omega_1x+\gamma_1),\\
&
\xi_1^{(3)}=T_2(x)=2\beta_2^T\sigma_2(\omega_2x)-1,
\end{align*}
from which we have
\begin{align*}
\bm{x}_1=
\begin{bmatrix}
\sigma_2(\omega_1x+\gamma_1)\\
\sigma_2(\omega_2x)
\end{bmatrix}
=\sigma_2(A_1x+b_1),
\quad \text{where} \quad
A_1=
\begin{bmatrix}
\omega_1\\
\omega_2
\end{bmatrix},\quad
b_1=
\begin{bmatrix}
\gamma_1\\
\bm{0}
\end{bmatrix},
\end{align*}
and
\begin{align*}
\bm{\xi}_{1} =
\begin{bmatrix}
 \xi_1^{(1)}\\
 \xi_1^{(2)}\\
 \xi_1^{(3)}
\end{bmatrix}
=A_{20}\bm{x}_1+b_{20},\quad
A_{20}=
\begin{bmatrix}
(c_1-c_3)\beta_1^T & \bm{0}\\
2c_3\beta_1 ^T & \bm{0}\\
\bm{0}         & 2\beta_2^T
\end{bmatrix},\quad
b_{20}=
\begin{bmatrix}
c_0\\
c_2\\
-1
\end{bmatrix}.
\end{align*}

\quad Immediate variables of the second hidden layer is (by Lemma \ref{lem:2})
\begin{equation}
\xi_2^{(1)}=\xi_1^{(1)}+\xi_1^{(3)}\xi_1^{(2)}
=\beta_1^T\sigma_2(\omega_1\xi_1^{(1)}+\gamma_1)
+\beta_1^T\sigma_2(\omega_1\xi_1^{(2)}+\gamma_1\xi_1^{(3)}),
\end{equation}
from which, we see the variable after the activations are
\begin{align*}
\bm{x}_2=
\begin{bmatrix}
\sigma_2(\omega_1\xi_1^{(1)}+\gamma_1)\\
\sigma_2(\omega_1\xi_1^{(2)}+\gamma_1\xi_1^{(3)})
\end{bmatrix}
=\sigma_2(A_{21}\bm{\xi}_{1} +b_{21}),
\quad
A_{21}=
\begin{bmatrix}
\omega_1 & \bm{0} & \bm{0}\\
\bm{0}   & \omega_1 & \gamma_1
\end{bmatrix},\
b_{21}=
\begin{bmatrix}
\gamma_1\\
0
\end{bmatrix}
\end{align*}
and
\begin{equation*}
A_2=A_{21}A_{20},\quad b_2=A_{21}b_{20}+b_{21}.
\end{equation*}
Noticing $p(x)=\xi_2^{(1)}$, so output of the neural network representing $p(x)$ is
\begin{equation*}
 p(x)=A_3\bm{x}_2+b_3,\quad \text{where}\quad
 A_3=[\beta_1^T,\beta_1^T],\quad b_3=0.
\end{equation*}

\item [(3)] For $n\ge 4$, let $m=\lfloor\log_2n\rfloor$,
then we can extend $p(x)$ as
\begin{equation}
p(x)=\sum\limits_{j=0}^{2^{m+1}-1}c_jT_j(x), \label{eq:chebexp_nge4}
\end{equation}
where $c_j=0$ if $n+1\le j\le 2^{m+1}-1$.
By Lemma \ref{lem:1D}, we can rewrite $p(x)$ as
\begin{align*}
p(x)
=\ & \left(c_0+\sum\limits_{j=1}^{2^m-1}c_jT_j(x)\right)
+ c_{2^m}T_{2^m}(x) +\sum\limits_{j=2^m+1}^{2^{m+1}-1}c_jT_j(x)\\
=\ & \left(c_0+\sum\limits_{j=1}^{2^m-1}(c_j-c_{2^{m+1}-j})T_j(x)\right)
+T_{2^m}(x)\left(c_{2^m} + 2\sum\limits_{j=1}^{2^m-1}c_{2^m+j}T_j(x) \right) \\
:=\ & r(x) +  T_{2^m}(x)q(x),
\end{align*}
where both $q(x)$ and $r(x)$ are polynomials of degree at most $2^m-1$.
If $q(x)$, $r(x)$, $T_{2^m}(x)$ are known, then by Lemma \ref{lem:2},
$p(x)=r(x)+q(x)T_{2^m}(x)$ can be realized by a $\sigma_2$ neural networks
with one-hidden layer of 8 $\sigma_2$ and 24 non-zero weights.
If both $r(x)$ and $q(x)$ can be realized in $m$ hidden layers, with no more
than $c\, 2^{m-1}-28$
nodes and non-zero weights, and $T_{2^{m}}(x)$ can be realized in $m$ hidden layers,
with $4m$ non-zero weights, which is true, then $p(x)$ can be realized
in  $m+1$ hidden layers, with no more than $c\,2^m - 28$ nodes and non-zero weights.
Here $c$ is a general constant that doesn't depend on $m$.
By induction, for any $n\ge 4$ the Chebyshev expansion Eq. \eqref{eq:chebexp_nge4}
can be realized in $\lfloor \log_2 n\rfloor + 1$ hidden layers, with
no more than $\mathcal{O}(n)$ neurons and total non-zero weights.
\end{itemize}
\end{proof}

The proof of Theorem \ref{thm:1D} presents a recursive ChebNet construction.
Since recursive algorithms are restricted in some computer languages. Next,
we give a non-recursive ChebNet construction by introducing a new set of
\emph{hierarchical Chebyshev bases}.

For $p(x) = \sum_{j=0}^n c_j T_j(x)$, $n\ge 2$, $m=\lfloor \log_2 n \rfloor$,
using Eq. \eqref{eq:cheb_dbl_rel}, we have
\begin{align}
p(x)
=\ & \left(c_0+\sum\limits_{j=1}^{2^m-1}(c_j-c_{2^{m+1}-j})T_j(x)\right)
+T_{2^m}(x)\left(c_{2^m}
+ 2\sum\limits_{j=1}^{2^m-1}c_{2^m+j}T_j(x) \right) \nonumber\\
=\ & \sum\limits_{j=0}^{2^m-1}\tilde{c}_j\hat{T}_j(x)
+ T_{2^m}(x) \left( \sum\limits_{j=0}^{2^m-1}\tilde{c}_{2^m+j}\hat{T}_j(x)
\right)\nonumber\\
=\ &\sum_{j=0}^n \tilde{c_j} \hat{T}_j(x), \label{eq:Relation-Basic}
\end{align}
where the \emph{hierarchical Chebyshev
basis} $\hat{T}_{2^m+l}(x)={T}_{2^m}(x)\hat{T}_l(x)$ for $l = 0, \ldots,
2^m-1$.
The definition of $\hat{T}_k(x)$ can be extended to all $k\in \mathbb{N}_0$ as
\begin{align}\label{eq:That}
\hat{T}_{n}(x)
=
\left\{
\begin{array}{ll}
T_{n}(x), & n={0,1,2,} \vspace{1ex}\\
T_{2^m}(x) \hat{T}_{n-2^m}(x), &  n\ge {3},
\end{array}
\right.
\end{align}
where $m=\lfloor \log_2 n \rfloor$. It is easy to see the transform between $\{c_j\}$ and $\{\tilde{c}_j\}$ is a linear transform. We denote it by
\begin{align} \label{eq:Sn}
\begin{bmatrix}
\tilde{c}_0 \\
\vdots \\
\tilde{c}_{{2^{m+1}-1}}
\end{bmatrix}
= S_m
\begin{bmatrix}
c_0\\
\vdots \\
c_{{2^{m+1}-1}}
\end{bmatrix}.
\end{align}
The transform matrix $S_{m}\!\in \mathbb{R}^{2^{m+1}\times 2^{m+1}} $ can be calculated using
following recursive formula
\begin{align} \label{eq:S_recur}
S_{j} =
(I_{2}\otimes S_{j-1})
\begin{bmatrix}
I_{2^j+1} &  -A^{(1)}_{j} \\
\bm{0} & 2I_{2^j-1}
\end{bmatrix}
,\ j\ge 1,\ \ \text{with}\
A^{(1)}_{j}
=
\begin{bmatrix}
\bm{0} \\
J_{2^{j}-1} \\
\bm{0}
\end{bmatrix} \in \mathbb{R}^{(2^{j}+1) \times (2^{j}-1) },
\end{align}
where $S_{0} = I_{2}$, and $I_{n}, J_{n}$ denotes the unit matrix of order $n$ and the oblique unit diagonal matrix of order $n$, respectively.

After transform from Chebyshev basis expansion $p(x)=\sum_{j=0}^n {c_j} {T}_j(x)$ to hierarchical Chebyshev basis expansion $p(x) = \sum_{j=0}^n \tilde{c_j} \hat{T}_j(x),\ p(x)$ can be realized exactly using algorithms almost identical to the ones building PowerNets proposed in \cite{li_better_2019}, the only difference is we need to replace the $\sigma_2$ network implementation of $x^{2^m} = (x^{2^{m-1}})^2$ in PowerNet with $T_{2^m}(x) = 2 T_{2^{m-1}}^2(x) - 1$.

Theorem \ref{thm:1D} can be extended to general RePU $\sigma_s, s > 2$
case, details are given in appendix. Note that, larger values of
  $s$ lead to $\sigma_s$ networks with higher order smoothness, which
  is good for problems with also high order smoothness. Moreover, the
  depth of network using larger $s$ needed to achieve certain order of
  convergence is smaller, since the maximum order can be achieved by a
  $L$-layer $\sigma_s$ network is $s^{L-1}$.  But the total number of
  nonzero weights of using $\sigma_s$ ChebNet to represent a given
  Chebyshev polynomial expansion increases linearly with $s$.  Please
  check the appendix for more details.

\subsection{ChebNets for multivariate Chebyshev polynomial expansions}
The construction described in last subsection, can be extended
to multivariate cases.

\subsubsection{Multivariate polynomials in tensor product and hyper-triangular space}
We first present the results of representing multivariate polynomials with fixed total degree.

\begin{theorem}\label{thm:d-dimension-GeneSpace}
If $p(\bm{x})$ is a multivariate polynomial with total degree $n$ in $d$
dimensions, then there exists a $\sigma_2$ neural network with $d\lfloor\log_2
n\rfloor+d$ hidden layers and no more than $\mathcal{O}(C^{n+d}_d)$ activation
functions and non-zero weights to {represent $p(\bm{x})$} exactly. {Here
$C_d^{n+d}$ is
the binomial coefficient.}
\end{theorem}
\begin{proof}
We first consider 2-d case.
Assume $p(x,y)=\sum_{i+j=0}^{n}c_{ij}T_i(x)T_j(y)$, $n\ge 3$. Let $m=\lfloor \log_2 n\rfloor$.
Since the transform from $\{c_j\}_{j=0}^n$ to $\{\tilde{c}_j\}_{j=0}^n$
using $S_{m+1}$ with zero paddings for $c_j, j=n+1, \ldots, 2^{m+1}-1$ terms will not lead to nonzero values of $\tilde{c}_j, j=n+1, \ldots, 2^{m+1}-1$,
the 2-d polynomial can be transformed into form
$p(x,y)=\sum_{i+j=0}^{n} \tilde{c}_{ij} \hat{T}_i(x) \hat{T}_j(y)$,
from which we have
\begin{equation}
p(x,y)
= \sum\limits_{i=0}^{n}\left(\sum\limits_{j=0}^{n-i} \tilde{c}_{ij} \hat{T}_j(y)\right)
\hat{T}_i(x)
=:\sum\limits_{i=0}^{n}B_i(y) \hat{T}_i(x),
\end{equation}
where $B_i(y)=\sum_{j=0}^{n-i} \tilde{c}_{ij} \hat{T}_j(y)$, $i=0,\ldots,n$.
That is, we regard $p(x,y)$ as a hierarchical Chebyshev polynomial expansion about variable $x$, which takes $B_i(y), i=0,\ldots,n$ as coefficients. It takes $3$ steps to construct a RePU network representing $p(x,y)$ exactly:
\begin{itemize}
\item[1)] For any $B_i(y), i=0,\ldots,n$, Theorem \ref{thm:1D} can be used to construct a $\sigma_2$ neural network to express $B_i(y)$ exactly. In other words,  there exists a $\sigma_2$ network $\Phi_1$ taking $y,x$ as input, $B_i(y),i=0,\ldots,n-1$, and $x$ as output. According to Theorem $\ref{thm:1D}$, depth of $\Phi_1$ is $\lfloor\log_2 n\rfloor +1$, number of activation functions and non-zero weights are both $\sum_{k=0}^{n-1}\mathcal{O}(k)=\mathcal{O}(\frac12 n^2)$. Since, the subnet built  for $B_i(y)$ according to Theorem \ref{thm:1D} have different depths, we need to
keep a record of $x$ at each layers by using \eqref{eq:ReQU_idx}. The cost for this purpose is $4 \lfloor \log_2 n \rfloor$, which is negligible comparing to $\mathcal{O}(\frac12 n^2)$.

\item[2)]
Taking $[B_0(y),\cdots,B_{n-1}(y),x]^T$ as input and combining the constructive process of Theorem $\ref{thm:1D}$, a neural $\sigma_2$ network $\Phi_2$ with $p(x,y)$ as output can be constructed. It is easy to see depth of $\Phi_2$ is $\lfloor\log_2 n\rfloor+1$, number of nodes and non-zero weights are both $\mathcal{O}(n)$.

\item[3)]
The concatenation of $\Phi_1$ and $\Phi_2$ is $\Phi$, which takes $y, x$ as input, outputs $p(x,y)$.  The details about concatenating two neural networks can be found in \cite{petersen_optimal_2018} or \cite{li_better_2019}. The depth of $\Phi$ is $2(\lfloor\log_2n\rfloor+1)$, number of non-zero weights and nodes are $\mathcal{O}(\frac12 n^2)$.
\end{itemize}
The $d>2$ cases can be proved using mathematical induction. Note that the
constant behind the big $\mathcal{O}$ can be made independent of dimension $d$
(without using zero padding, but the proof is a little bit lengthy).
For the case $n \le 2$, one can build neural networks with only one hidden
layer (see, e.g. \cite{mhaskar_approximation_1993}).
 The theorem is proved.
\end{proof}

Using similar approach, we can construct optimal ChebNet for
polynomials in tensor product space $Q_N^d(I_1\times \cdots \times
I_d):=P_N(I_1)\otimes\cdots\otimes P_N(I_d)$, the result is given in
the following theorem.
\begin{theorem}
Polynomials from a tensor product space $Q_N^d(I_1\times \cdots \times I_d)$ can be realized without error with a deep $\sigma_2$ neural network, in which the depth is $d\lfloor\log_2N\rfloor+d$, the numbers of activation functions and non-zero weights are no more than $\mathcal{O}(N^d)$.
\end{theorem}

\subsubsection{Multivariate polynomials in sparse downward closed polynomial spaces}

For high dimensional problems, it is obvious that the degree of freedoms increases
exponentially as dimension $d$ increases if tensor-product or similar grids are used,
which is known as curse of dimensionality.
Fortunately, a lot of practical
high-dimensional problems have low intrinsic dimensions, see e.g. \cite{wang_why_2005}, \cite{Yserentant2004}.
A particular example is the class of
high-dimensional smooth functions with bounded mixed
derivatives, for which sparse grid (or hyperbolic cross) approximation is a very popular approximation tool
(see e.g. \cite{smolyak_quadrature_1963}, \cite{bungartz_sparse_2004}). In
the past few decades, sparse grid method and hyperbolic
cross approximations have found many applications, such as
general function approximation
\citep{barthelmann_high_2000, shen_sparse_2010,shen_approximations_2014},
solving partial differential equations (PDE)
\citep{bungartz_adaptive_1992, lin_sparse_2001, shen_efficient_2010, shen_efficient_2012, guo_sparse_2016, rong_nodal_2017},
computational chemistry \cite{griebel_sparse_2007, avila_solving_2013, shen_efficient_2016},
uncertainty quantification
\cite{schwab_sparse_2003a, nobile_sparse_2008},
etc.

The sparse grid finite element approximation was recently used by
\cite{montanelli_deep_2017} to construct
a new upper error bound for deep ReLU network approximations in
high dimensions. Optimal deep RePU networks
based on sparse grid and hyperbolic cross spectral approximations are constructed by \cite{li_better_2019},
which give better approximation bounds for sufficient smooth functions.

Now, we describe how to construct Deep RePU networks
based on high dimensional sparse Chebyshev polynomial approximations without transform into power series form as done in \cite{li_better_2019}.
Both hyperbolic cross set and sparse grids belong to a more general set:
downward closed sets (see e.g.
\citep{cohen_approximation_2015}\citep{li_better_2019}),
which is defined below. So we only present the result
for downward closed polynomial spaces here.

\begin{Definition}
A linear polynomial space $P_C$ is said to be {\em downward closed},
if it satisfies the following:
\begin{itemize}
	\item if $d$-dimensional polynomial $
	p(\bx) \in P_{C}$, then $ \partial^{\bm{k}}_{\bx} p(\bx)
	\in P_{C}$ for any $\bm{k}\in \bbN_0^d$,
	\item there exists a set of bases that consist of monomials
	only.
\end{itemize}
\end{Definition}

Now we give a conclusion on approximating Chebyshev polynomial expansions in
downward closed polynomial space $P_C$.

\begin{theorem}\label{thm:dwclosed}
Let $p(\bm{x})$,$\bm{x}\in \mathbb{R}^d$ be a polynomial in downward closed polynomial space $P_C$.
Let $n$ be the dimension of $P_C$. Then there exists a $\sigma_2$ neural network with no more than $\sum_{i=1}^{d}\lfloor\log_2N_i\rfloor+d$ hidden layers, $\mathcal{O}(n)$ activation functions and non-zero weights, can represent $p$ exactly, where $N_i$ is the maximum polynomial degree in $x_i$ for functions in $P_C$.
\end{theorem}

The proof is similar to Theorem \ref{thm:d-dimension-GeneSpace}.
One key fact is the transforms from expansions using standard Chebyshev polynomials $T_k \in P_C$ as bases to expansions using hierarchical Chebyshev polynomials $\hat{T}_k$ as bases do not add nonzero coefficients for the bases $\hat{T}_k \notin P_C$. After rewriting $p(\bm{x})\in P_C$ into linear combinations of $\hat{T}_k$ basis,
one can construct corresponding deep RePU networks by dimension induction similar as in
Theorem \ref{thm:d-dimension-GeneSpace} or procedure described in Theorem 4.1 of \cite{li_better_2019}.

\begin{remark}\label{rmk:Compare}
The structures of deep RePU networks constructed by using hierarchical Chebyshev
polynomial expansion (i.e. ChebNet) and those constructed by using power
series expansions (i.e. PowerNet)
are very similar. There is one small difference. In ChebNet, to calculate
$T_{2^{m+1}}$ from $T_{2^m}$, a constant shift vector 1 is added comparing to
calculating $x^{2^{m+1}}$ from $x^{2^m}$ in PowerNet.
So the depth and network complexity of ChebNet and PowerNet are exactly the
same.
\end{remark}

\begin{remark}\label{rmk:sparsegridhyperbolic}
	Since hyperbolic cross and sparse grid polynomial spaces are
	special cases of downward closed polynomial spaces, the Theorem \ref{thm:dwclosed} can be directly applied to sparse grid and hyperbolic cross
	polynomial spaces.
\end{remark}

\section{Approximating general smooth functions}

It is well known that polynomial approximation converge very fast
for approximating smooth functions.
The ChebNets constructed in last section can be used to approximate
general smooth functions.
There are three steps in using ChebNets for approximating general smooth
functions.
\begin{enumerate}
	\item Construct the Chebyshev polynomial approximation for given smooth function. This can be done by using fast Fourier transform \citep{trefethen_spectral_2000}. For low dimensional problem, e.g. for $d<4$, one may use tensor-product grids. For high dimensional
	problem, one may use sparse grids or other downward closed sparse polynomial approximations. The fast Chebyshev transform on sparse grids is constructed by \cite{shen_efficient_2010}. For problems in unbounded high dimensional domain,
	one can use mapped Chebyshev method, fast convergence \citep{shen_approximations_2014} and fast Chebyshev transform \citep{shen_efficient_2012} are also available.

	\item Construct the corresponding ChebNets for the polynomial
	approximation
	obtained from the first step using constructions described in Section 2.

	\item Fine-tune the constructed ChebNets using more data to improve the
	approximation
	accuracy.
\end{enumerate}

There are several remarks on the approximation properties of the ChebNet for
general
smooth functions.

\begin{remark}
	{If the coefficients of the truncated Chebyshev expansion and the coefficients of truncated power series are calculated
		without any numerical error, then the approximation property
		of Theorem \ref{thm:MdSobolev} applies to
		both PowerNets and ChebNets.
	However, in practice, we work with float-point numbers with finite
	precision, where ChebNets should be more stable. The
conclusion comes from equation \eqref{eq:condnum_err} and Table
\ref{ tbl:2} in next section, where we see the condition number of
the transform matrix associated PowerNets grows geometrically, while it grows
linearly for ChebNets.}
\end{remark}


\begin{remark}
	{
	When the function values are given on Chebyshev-Gauss-Lobatto
	points, we can compute the Chebyshev coefficients stably using $\mathcal{O}(n\log n)$ operations, here $n$ is the number of points. On the other hand, the computing of power series needs
    $\mathcal{O}(n^2)$ operations, and some significant digits may be lost due to the bad condition number.
	When the function values are given on random points, then we may
    calculate Chebyshev coefficients and power series using the method of least squares.
		The ReLU networks based on polynomial approximations have a similar computational cost for
	construction, but the resulting networks are much larger.}
\end{remark}

\begin{remark}
	For high dimensional problems, we use sparse grid Chebyshev approximations
	to build ChebNets. It is known that sparse grid is not isotropic, i.e.
	using different coordinates may have different convergence properties.
	Another issue is
	that the complexity of sparse grids still weak-exponentially depends on
	dimension $d$. To overcome these two issues, one may add one extra
	full connected RePU subnet to accomplish dimension reduction or
	coordinates transform before feeding the data into ChebNets. The networks
	can be trained separately or collectively.
\end{remark}

\begin{remark}
    Regarding the question whether ChebNets can approximate any smooth
    function, we quote the following proposition, which is a corollary
    of the Stone approximation theorem, see e.g. Chapter 6 of
    \cite{cheney_introduction_1982}.
    \begin{proposition}
        Let $X$ be a compact set in $n$ dimensional space. The
        polynomials in $n$ variables form a dense set in $C(X)$.
    \end{proposition}
    So, provided that best polynomial approximations can be obtained,
    we can construct corresponding ChebNets to approximate any smooth
    functions of interested defined on a compact set.  But best polynomial approximations are
    not easy to obtain. We usually use interpolation or spectral
    transform to obtain polynomial approximations of smooth functions,
    which in general are not but close to the optimal ones.  ChebNets
    constructed basing on these polynomial approximations can be
    further fine-tuned to get more and more close to optimal, which
    will be demonstrated in next section.
\end{remark}

\section{Numerical experiments}


In this section, we show the performance of ChebNets in approximating given
smooth functions, and compare the results with PowerNets.
In this paper we focus on the performance differences between ChebNets and
PowerNets,
so we only use 1-dimensional examples.

\subsection{Numerical results}

We test two smooth functions defined as follows.
\begin{itemize}
    \item[(1)]Gauss function:
    \begin{equation}
    {f_1}(x)=\exp(-x^2),\quad x\in[-1,1]. \label{eq:Gauss}
    \end{equation}
    \item[(2)] A {smooth but not analytic} function:
    \begin{align}
    {f_2}(x)=\left\{
    \begin{array}{ll}
    \exp(-\frac{1}{x^2}), & x\neq 0,\\ 0, & x=0.\\
    \end{array}
    \right. \label{eq:Cauchy_fun}
    \end{align}
\end{itemize}

We use truncated $N$ item  Legendre and Chebyshev polynomial approximations for
PowerNet and ChebNet, respectively. For PowerNet, we transform the Legendre
approximation to power series representation before using the construction
method proposed in \cite{li_better_2019}.

The $L^2$ errors of ChebNets and PowerNets in approximating
  function $f_2$ are presented in Table \ref{ tbl:1}, from which we see
  that PowerNets give slightly better results than ChebNets when $n\le
  24$, which is because the underlying Legendre polynomial expansions
  used by PowerNet are orthogonal in standard $L^2$ space, while Chebyshev
  polynomials are orthogonal with respect to weighted $L^2$ inner
  product. But, for $n \ge 32$, the performance of PowerNets
  deteriorate quickly as $n$ increases, while no observable
  deterioration happens to ChebNets. 
  \rev{This is due to the bad-condition problem associated with monomial bases, 
which will be explained in next subsection and Table \ref{ tbl:2}. 
We note that, if high-accuracy is not required, then PowerNets could give slightly 
better results than ChebNets, as showed in Table \ref{ tbl:1}. 
But for problems in which accuracy and efficiency are important, ChebNets 
with high-degree polynomial approximations are a better choice.}

\begin{table}[h]
    \centering
    \caption{\label{ tbl:1} {The $L^2$ error of using PowerNets and
        ChebNets to approximate function $f_2$ defined in
        \eqref{eq:Cauchy_fun} using different numbers of
        polynomials.}}
    \begin{tabular}{rll}
        \hline
        degree ${n}$ & PowerNet &  ChebNet \\
        \hline
        8  & 3.98e-3 & 4.13e-3\\
        16 & 3.94e-4 & 4.14e-4\\
        24 & 5.38e-5 & 5.57e-5\\
        32 & 1.57e-5 & 9.18e-6\\
        36 & 4.78e-2 & 2.88e-6\\
        40 & 1.78e5 & 1.46e-6 \\
        44 & 5.53e10 & 8.41e-7\\
        \hline
    \end{tabular}
\end{table}

 Next, we use $200$ uniform points from the interval $[-1,1]$ as training
 data to fine-tune PowerNets and ChebNets. All the experiments are
 executed using Tensorflow with RMSPropOptimizer, where $\gamma=0.99$,
 $\eta=0.00001$.
The loss function during training is the average of $l_2$ norm squared.

%

\begin{figure}[!thbp]
\centering
\includegraphics[width=0.48\textwidth]{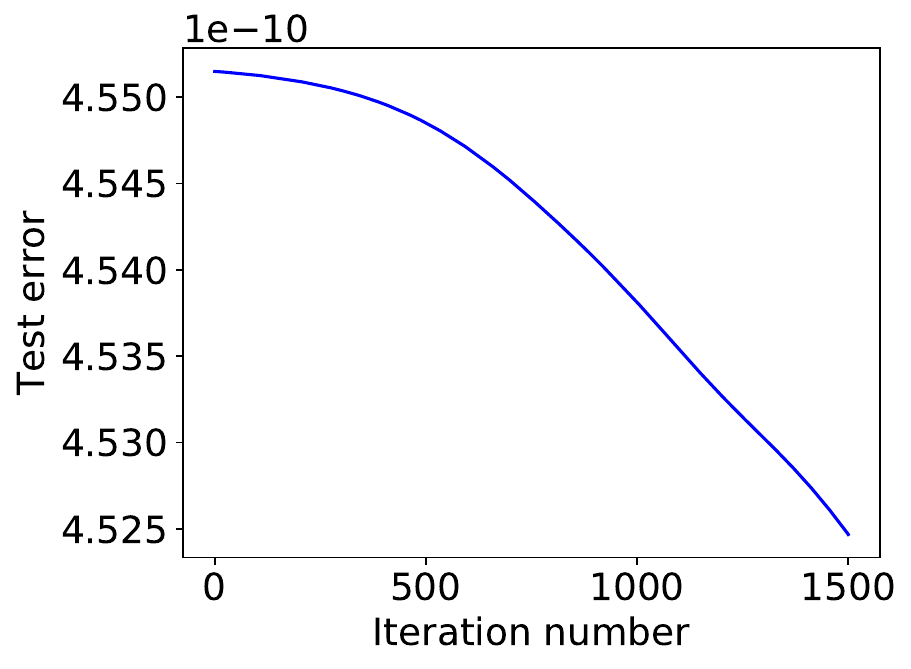}
\includegraphics[width=0.48\textwidth]{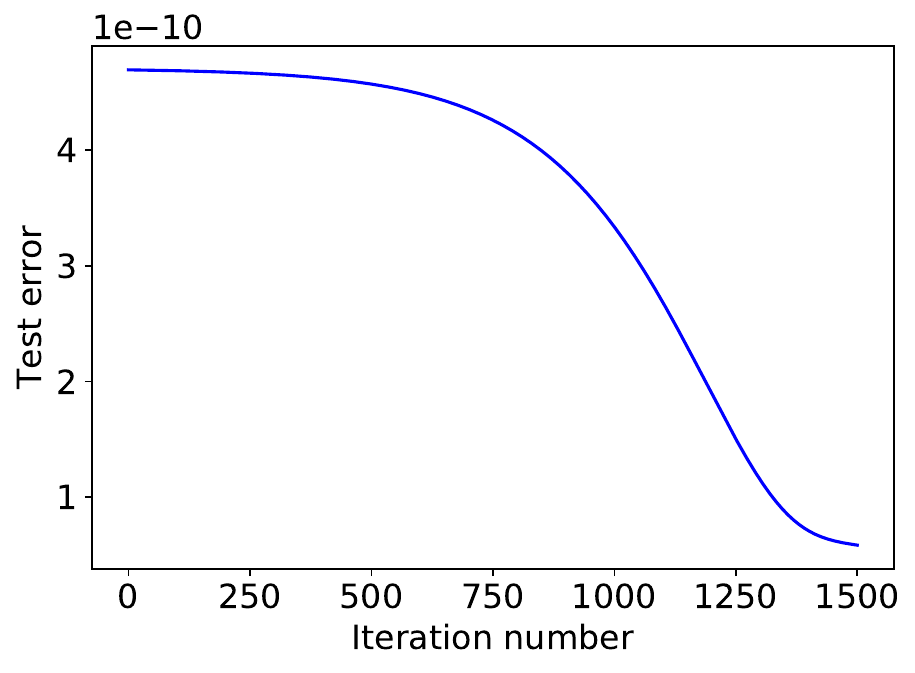}
\caption{
	Training results of PowerNet (left) and ChebNet (right)
    constructed to approximate Gauss
	function
	with ${n}=15$. {$L^2$ norm is used for test error.}}
\label{fig:PvC_GuassN15}
\end{figure}

We take polynomial approximations of degree ${n}=15$ for Gauss function defined
in \eqref{eq:Gauss} and show the training
performances in Figure \ref{fig:PvC_GuassN15}, where the horizontal axis
represents the iteration number of training, and the vertical axis is the error
on training set. We see the initial errors are almost the same. After training,
the error of ChebNet on the right plot is reduced to less than $1/6$ of the
original error, while the error of PowerNet decreases only a very small
percentage.

\begin{figure}[!thbp]
\centering
\includegraphics[width=0.48\textwidth]{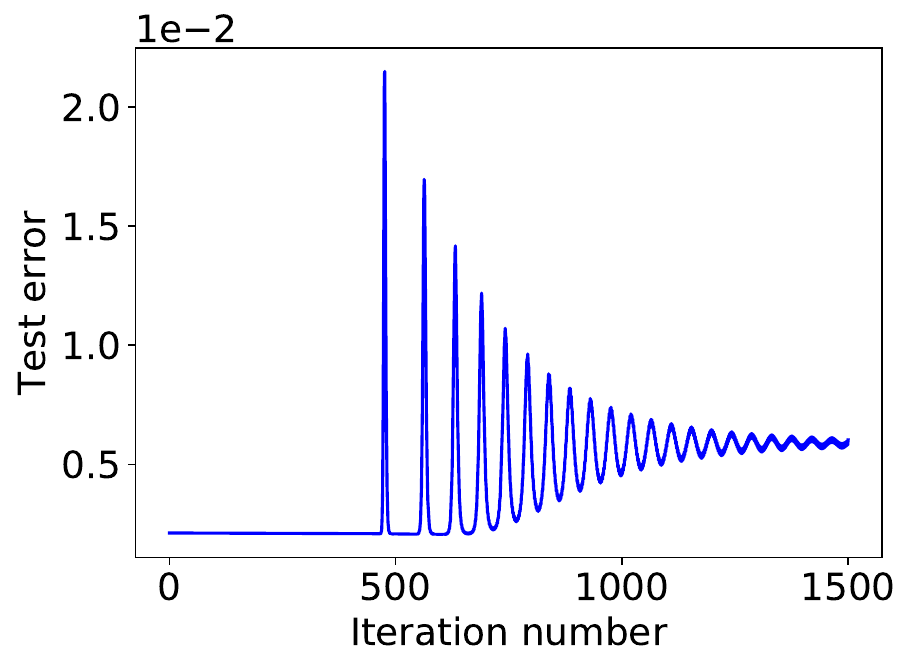}
\includegraphics[width=0.45\textwidth]{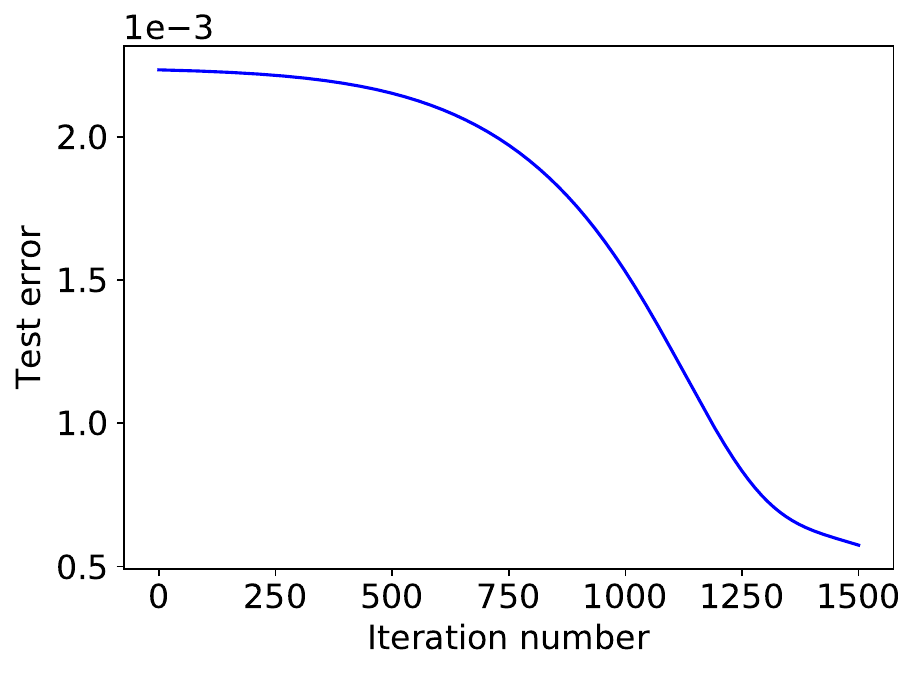}
\caption{\label{fig:PvC_CauchyN11}Training results of PowerNet (left) and
ChebNet (right) constructed to approximate function $f_2$ defined in
\eqref{eq:Cauchy_fun} with
${n}=11$.}
\end{figure}

Polynomial approximation of degree $n=11$ is taken for $f_2$ defined in
\eqref{eq:Cauchy_fun} and the results are shown in Figure
\ref{fig:PvC_CauchyN11}.
Similar to Gauss function, the error of ChebNet is reduced much more than
PowerNet after training. Actually, in this case the PowerNet is very hard to
train, the error increased after training. Notice that the precision is not
high by taking ${n}=11$, so we hope to achieve high-accuracy and compare the
corresponding results for ${n}=30$. However, PowerNet blows up after the $1$st
iteration of training for ${n}=30$. On the other hand side, the ChebNet can
still be trained and obtains better accuracy.


\subsection{Explanations of the numerical experiments}

We give some explanations on the numerical results here.
To approximate a function $f(x)$, we do in PowerNet as follows:
\begin{equation}
f(x)\approx f_n=\sum\limits_{j=0}^{n}
a_jL_j(x)=\sum\limits_{j=0}^{n}\tilde{a}_jx^j,
\end{equation}
where { $L_j(x)$  are Legendre polynomials,}
$\tilde{a}_j$ are calculated using a linear transform
\begin{align}
\left[
\begin{array}{c}
\tilde{a}_0\\
\tilde{a}_1\\
\vdots\\
\tilde{a}_n
\end{array}
\right]
=B_n  \left[
\begin{array}{c}
a_0\\
a_1\\
\vdots\\
a_n
\end{array}
\right].
\end{align}

Correspondingly, the following formula are used in constructing ChebNet:
\begin{equation}
f(x)\approx f_n=\sum\limits_{j=0}^{n}c_jT_j(x)=\sum\limits_{j=0}^{{n} } \tilde{c}_j\hat{T}_j(x),
\end{equation}
where $c_j$ and $\tilde{c}_j$ satisfy
\begin{align}
\left[
 \begin{array}{c}
 \tilde{c}_0\\
 \tilde{c}_1\\
 \vdots\\
 \tilde{c}_{n}
 \end{array}
\right]
=H_{n} \left[
\begin{array}{c}
 c_0\\
 c_1\\
 \vdots\\
 c_{n}
\end{array}
\right].
\end{align}
Here $H_{n} $ is the first $({n} +1)\times ({n} +1)$ sub-matrix of $S_m$ with $m=\lfloor \log_2 {n} \rfloor$.

\begin{figure}[!thbp]
    \centering
    \includegraphics[width=0.45\textwidth]{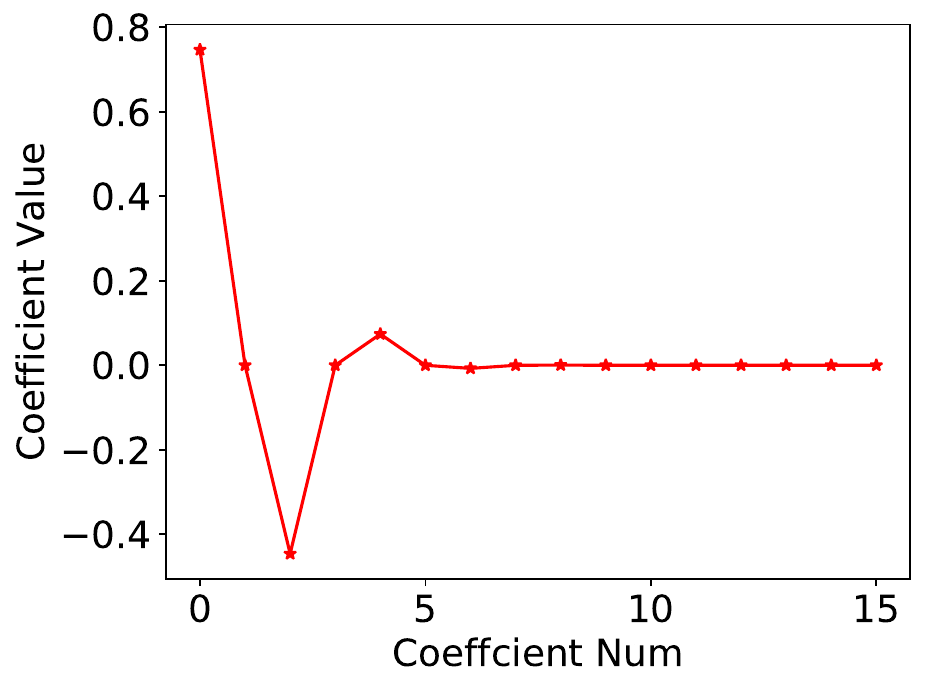}
    \includegraphics[width=0.45\textwidth]{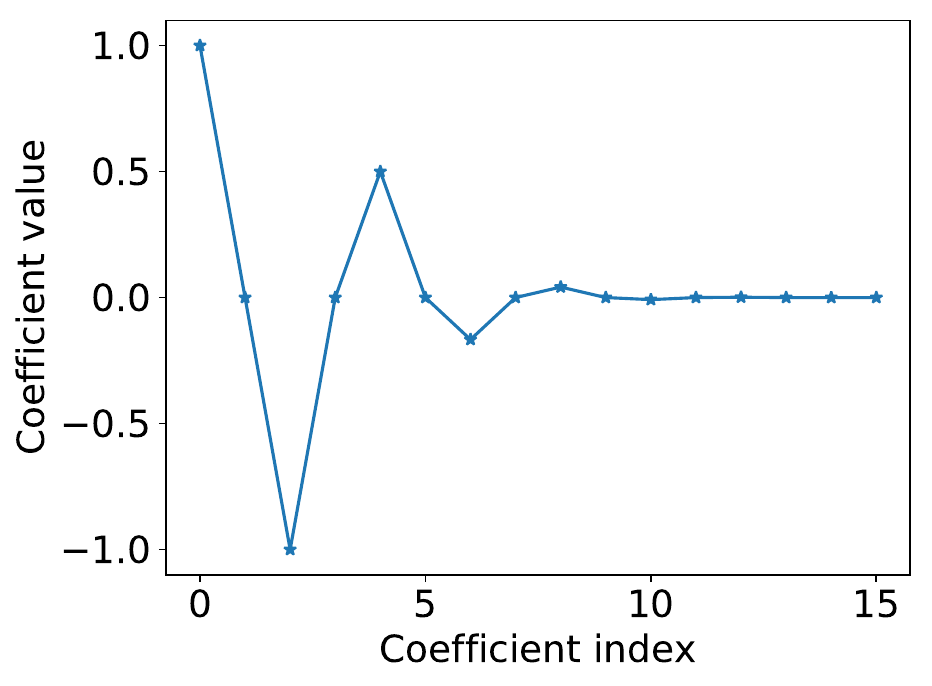}
      \caption{The coefficients of Legendre expansion: $a_j, j=0,\ldots,{n} $
      (Left)  and power series expansion: $\tilde{a}_j, j=0,\ldots,{n} $
      (Right) for Gauss function with ${n} =15$.}
    \label{fig:PowerCoefN15}
\end{figure}

\begin{figure}[!thbp]
    \centering
    \includegraphics[width=0.45\textwidth]{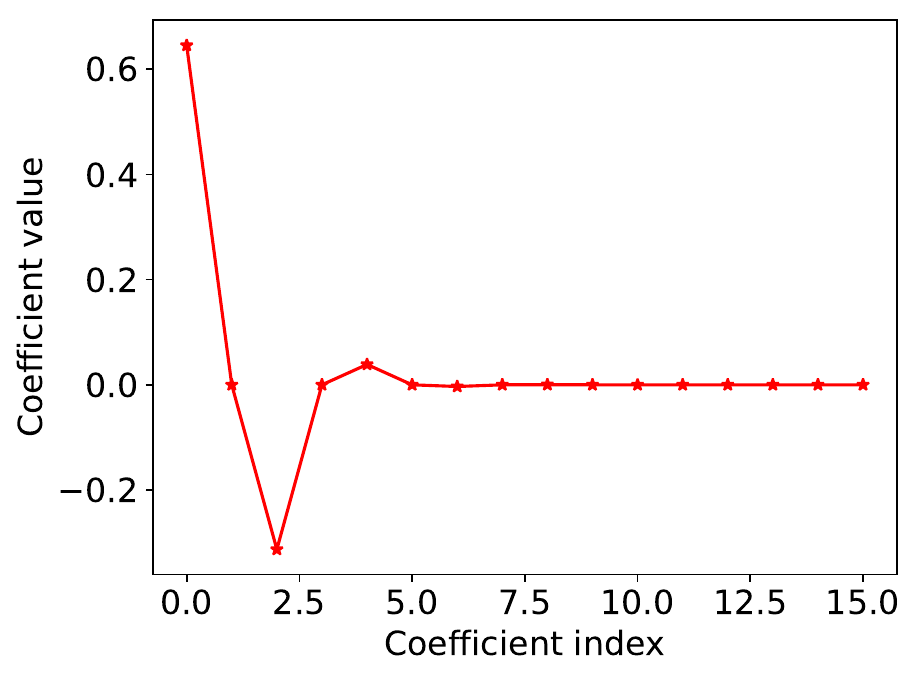}
    \includegraphics[width=0.45\textwidth]{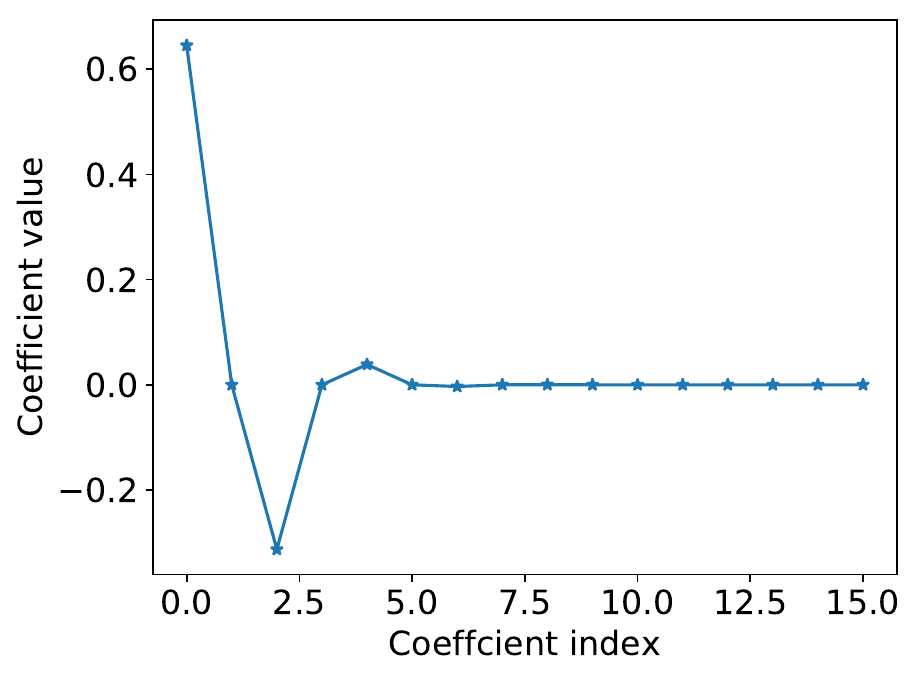}
    \caption{The coefficients of Chebyshev expansion: $c_j, j=0,\ldots,{n} $
    (Left)  and coefficients of hierarchical Chebyshev expansion: $\tilde{c}_j,
    j=0,\ldots,{n} $ (Right) for Gauss function with ${n} =15$.}
    \label{fig:ChebCoefN15}
\end{figure}

\begin{figure}[!thbp]
   \centering
    \includegraphics[width=0.45\textwidth]{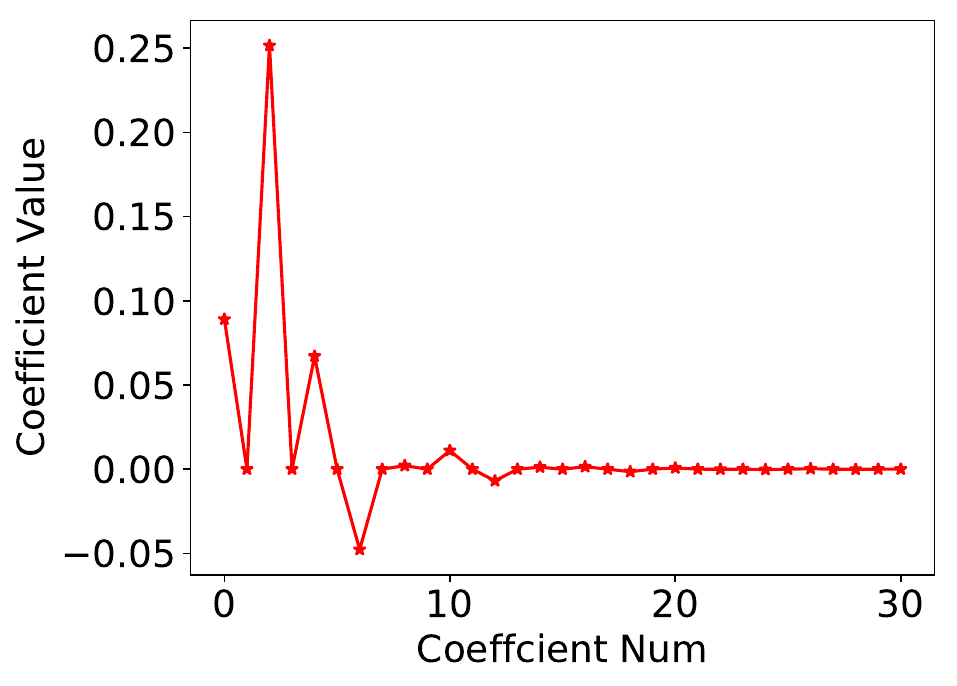}
    \includegraphics[width=0.48\textwidth]{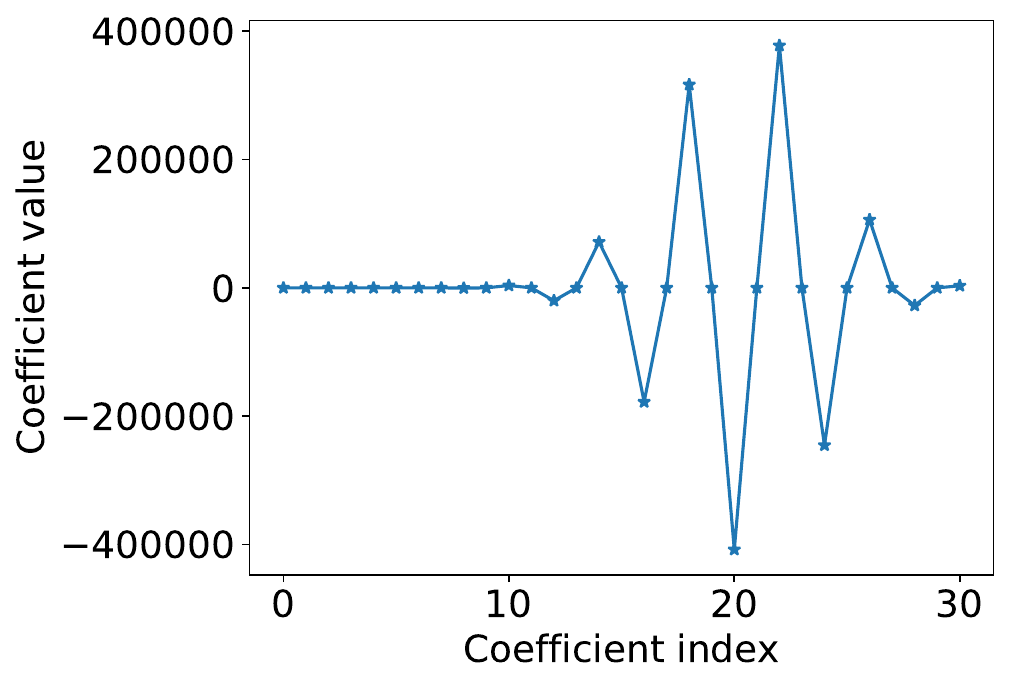}
  \caption{The coefficients of Legendre expansion: $a_j, j=0,\ldots,{n} $
  (Left)  and coefficients of power series expansion: $\tilde{a}_j,
  j=0,\ldots,{n} $ (Right) for function $f_2$ defined in \eqref{eq:Cauchy_fun}
  with ${n} =30$.}
  \label{fig:PowerCoefN30}
\end{figure}

\begin{figure}[!thbp]
   \centering
 \includegraphics[width=0.45\textwidth]{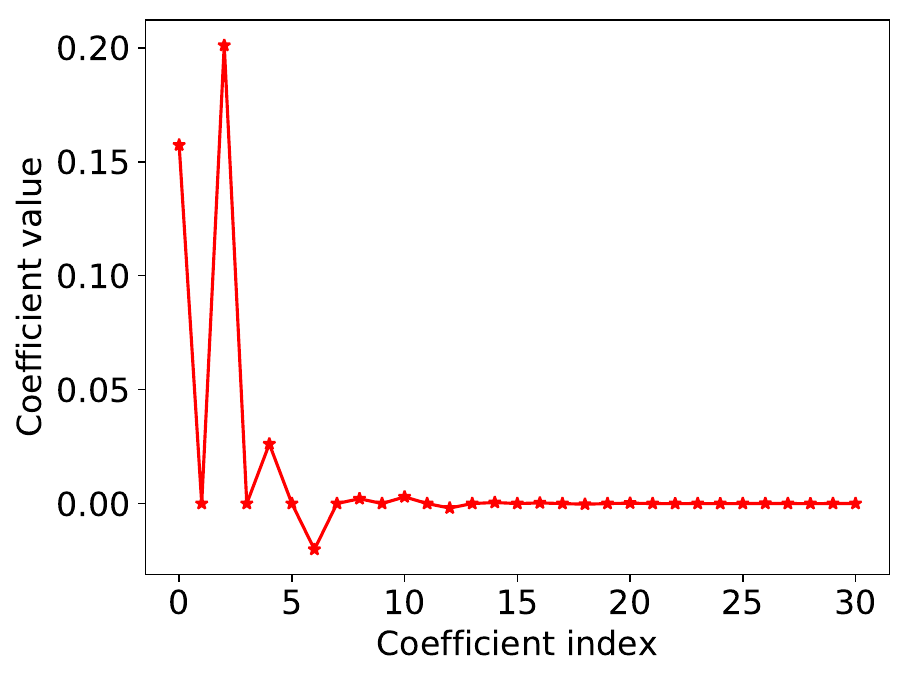}
  \includegraphics[width=0.45\textwidth]{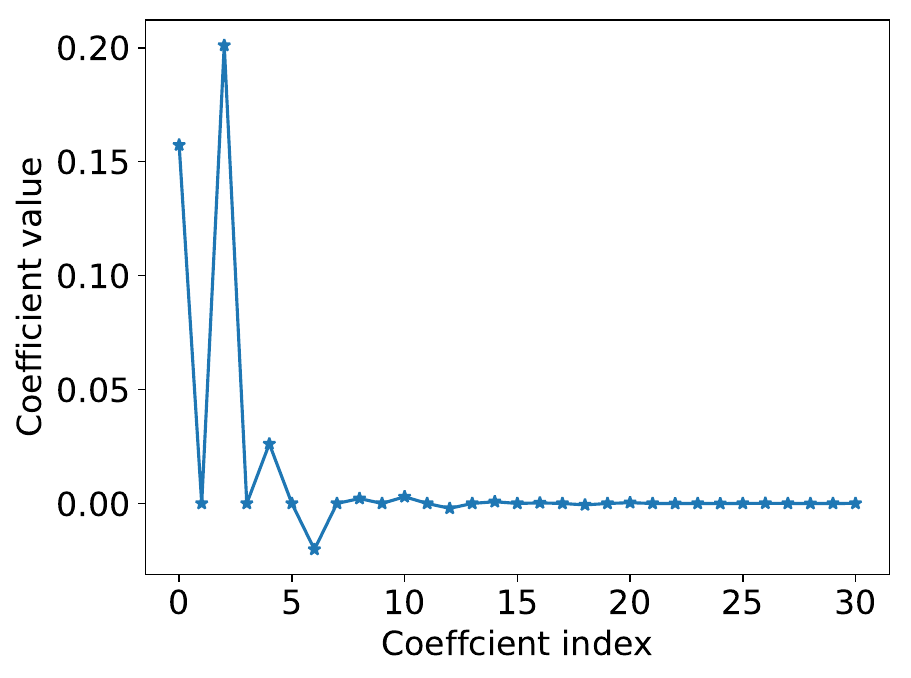}
    \caption{The coefficients of Chebyshev expansion: $c_j, j=0,\ldots,{n} $
    (Left),  and coefficients of hierarchical Chebyshev expansion:
    $\tilde{c}_j, j=0,\ldots,{n} $ (Right), for function $f_2$ defined in
    \eqref{eq:Cauchy_fun} with ${n} =30$.}
    \label{fig:ChebCoefN30}
\end{figure}

Now we plot $c_j,\tilde{c}_j,a_j,\tilde{a}_j$, $j=0,\ldots,{n} $ in Figure
\ref{fig:PowerCoefN15}-\ref{fig:ChebCoefN30} for approximating Gauss function
$f_1(x)$ using $15$ terms and for approximating function $f_2$ with $30$ terms.
From these figures, we see the differences between $c_j$ and $\tilde{c}_j$ are
small, but the differences between $a_j$ and $\tilde{a}_j$ are very large,
when ${n} $ is large.
Especially, in approximating function $f_2$ with ${n} =30$ using power series expansion, we have some coefficients almost as large as $4\times10^{5}$, see Figure \ref{fig:PowerCoefN30}. The big coefficients make the resulting PowerNets deteriorate and hard to train.



To explain why big coefficients happen, we calculate the condition numbers of $B_{n} $ and $H_{n} $, denoted by $\kappa(B_{n} )$ and $\kappa(H_{n} )$, and the results are showed in Table \ref{ tbl:2}. We see from the table that the condition number of $\kappa(B_{n} )$ increases exponentially fast, but the condition number of $\kappa(H_{n} )$ increases slowly as ${n} $ increases. The large condition number of $B_{n} $ indicates that
transform from Legendre expansion to power series is not a good approach,
which may introduce large numerical error due to the large condition number.
Meanwhile, the large power series expansion coefficients embedded
into PowerNets make the nets hard to train using gradient method, due to
large variation between gradients to different internal network parameters.

In Table \ref{tbl:cond-s}, we also show the condition numbers
of the corresponding transforms $H_{n} $, associated to the $\sigma_s, s=2,3,4,5,6$
network construction based on Chebyshev approximations.
From this table, we see the condition number of the transform is not
very sensitive to $s$.

\begin{table}[h]
	    \centering
	    \caption{\label{ tbl:2}The condition number of $B_{n} $, $H_{n} $ for $s=2$.}
	    \begin{tabular}{lcccc}
		\hline
		${n} $ & $10$ &  $20$ & $30$  & $40$\\
		\hline
		$\kappa(B_{n} )$& $8.750e2$ & $4.1e6$& $2.2e10$& $1.3e14$\\
		$\kappa(H_{n} )$& $1.234e1$  & $2.555e1$& $2.555e1$ & $5.228e1$\\
		\hline
	    \end{tabular}
\end{table}

\begin{table}[!thbp]
\centering
\caption{\label{tbl:cond-s} The condition number of $H_{{n} }$ for different $s$.}
\begin{tabular}{c|ccccccc}
\hline
{${n} $} & $10$ &  $50$ & $100$  & $200$ & $500$ & $1000$ & $2000$\\
\hline
$s=2$& $1.234e1$ & $5.228e1$ & $1.062e2$& $2.150e2$ & $4.338e2$ & $8.736e2$ & $1.757e3$\\
$s=3$& $1.371e1$  & $4.432e1$& $1.415e2$ & $1.415e2$ & $4.493e2$ & $1.422e3$ & $1.422e3$ \\
$s=4$& $6.008$  & $3.285e1$& $1.772e2$ & $1.772e2$ & $9.538e2$ & $9.538e2$ & $5.130e3$ \\
$s=5$& $9.1002$  & $6.582e1$ & $6.582e1$ & $4.714e2$ & $4.714e2$ & $3.371e3$ & $3.371e3$ \\
$s=6$& $1.22e1$  & $1.102e2$ & $1.102e2$ & $1.102e2$ & $1.029e3$ & $1.029e3$ & $9.376e3$\\
\hline
\end{tabular}
\end{table}

\section{Conclusion}
In this paper, we improve the RePU network construction based on polynomial approximation proposed recently in \cite{li_better_2019}.
By removing the procedure of transforming a polynomial into a power series, and
constructing
deep RePU networks directly from Chebyshev polynomial approximation, we eliminate the numerical instability associated with the non-orthogonal monomial bases.
Due to the availability of fast Chebyshev transforms,
the proposed new approach, which we call ChebNets can be efficiently
applied to a large class of smooth functions.
Considering other good properties that RePU networks have: 1) obtain
high order convergence with less layers than ReLU networks; 2) fit in the
situation where derivatives are involved in the loss function; we expect that
deep RePU networks, and particularly ChebNets, be more
efficient in applications where high dimensional functions to be approximated are  smooth.

\section*{Acknowledgments}
This work was partially supported by
 NNSFC under Grant No. 12171467, 11771439, 12161141017.

\section*{Statements and Declarations}
The authors declare that they have no conflict of interest.

\begin{appendices}

\section{Realization of univariate Chebyshev polynomials with general
RePUs}

Here, we show how to extend the method in Theorem \ref{thm:1D} to general
RePU $\sigma_{s}(x),\ (s=2,3,\ldots)$. The main results is given in the following theorem.

\begin{theorem} \label{thm:1d-s}
	Assume $p(x)=\sum_{j=0}^{n}b_jT_j(x) \ (b_n\neq 0)$, $x\!\in\! \mathbb{R}$, then there exists a $\sigma_s$ neural network with at most $\lceil\log_s n\rceil$+1 hidden layers to represent $p(x)$ exactly. The numbers of the total neurons and non-zero weights are {of order
	$\mathcal{O}(n)$ and $\mathcal{O}(sn)$, respectively}.
\end{theorem}

To prove this theorem, we need following two lemmas.

\begin{lemma}\label{lem:cheb_expan}
	For Chebyshev polynomials of the first kind $T_{n}(x)$, we have the following relationship
	{\begin{align}
		T_{rs+k}(x)
		= 2 \sum^{r}\limits_{i=1}\delta^{-}_{r-i}\left[T_{is}(x) T_{k}(x) - T_{(i-1)s}(x) T_{s-k}(x) \right]
		+ \delta^{+}_{r} T_{s-k}(x)
		+ \delta^{-}_{r} T_{k}(x),  \label{eq:cheb_exp}
	\end{align}
where $2\le s\in \mathbb{N}$, $r\in \mathbb{N}$, $k=1,2,\ldots,s-1$, and $\delta^{\mp}_{r}:=\frac{1\pm(-1)^{r}}{2}$. Note that $\delta^{\mp}_r = 1$ means $r$ is an even/odd number.}
\end{lemma}

\begin{lemma}\label{lem:cheb_compon}
For a polynomial $P^{(k+1)}(x)\!=\!\sum^{s^{k+1}-1}_{j=0}b_{j}T_{j}(x)$ of degree  $s^{k+1}-1 $, there exists  a set of polynomials $\{\widetilde{P}^{(k)}_{l}(x)\}^{s-1}_{l=0}$ of degree up to $s^{k}-1 $, such that
\begin{align}\label{eq:poly_recur_s2}
P^{(k+1)}(x)
= \widetilde{P}^{(k)}_{0}(x) +
\widetilde{P}^{(k)}_{1}(x) {T_1}\Big(T_{s^{k}}(x) \Big)+
\cdots +
\widetilde{P}^{(k)}_{s-1}(x) {T_{s-1}}\Big(T_{s^{k}}(x) \Big),
\end{align}
where
{
\begin{align}
&&&\widetilde{P}^{(k)}_{l}(x)
=
\sum^{s^{k}-1}\limits_{j=0}\widetilde{A}^{(k)}_{l,j} T_{j}(x),
\quad l=0,1,...,s-1, \nonumber\\
&&\widetilde{A}^{(k)}_{l,0}
=\
e^{(l \cdot s^{k})}_{s,k} \cdot \bm{b}_{k}, \quad
&\widetilde{A}^{(k)}_{l,j}
=\  (2-\delta_{l0}) \sum^{s-1}\limits_{r=l}\Big(\delta^{-}_{r-l}e^{(r
s^{k}+j)}_{s,k} - \delta^{+}_{r-l}e^{((r+1) s^{k}-j)}_{s,k}\Big)\cdot
\bm{b}_{k}, 
\nonumber\\
& & &e^{(i)}_{s,k} = [0,\cdots,0,\overset{\overset{\overset{(i)}{\uparrow}}{
}}{1},0\cdots,0]\in \mathbb{R}^{1\times s^{k+1}}.\nonumber
\end{align}
In above expressions, $\delta_{nm}$ is the Kronecker delta, $l=0,1,\ldots,s-1$,
$j=1,2,\ldots,s^{k}-1$, and $\bm{b}_{k}=[b_{0},\ldots,b_{s^{k+1}-1}]^{T}$.
}
\end{lemma}

\begin{proof}[Proof of Lemma \ref{lem:cheb_expan}.]
	{For $r=1$, Eq \eqref{eq:cheb_exp} is reduced to
	\begin{align}
	T_{s+k}(x) = 2T_{s}(x)T_{k}(x)- T_{s-k}(x), \label{eq:r1}
	\end{align}
	which is Eq \eqref{eq:cheb_dbl_rel} in Lemma \ref{lem:1D}.

	When $r\geq 2$, for $2\le j \le r$, using Eq \eqref{eq:cheb_dbl_rel} twice, we have
	\begin{align}
		T_{js+k}(x)
		=\:
		2T_{js}(x) T_{k}(x) -
		2T_{(j-1)s}(x)T_{s-k}(x) + T_{(j-2)s+k}(x),
		\quad k = 1,2,..., s-1.
		\label{eq:rj}
	\end{align}
If $r \ge 2$ is an odd number,
summing up \eqref{eq:rj} for $j=r, r-2, \ldots, 3$ and \eqref{eq:r1}, we obtain
\begin{align}
T_{rs+k}(x) = 2\sum_{j=3, \text{odd}}^r \big[ T_{js}(x)T_k(x) - T_{(j-1)s}(x)T_{s-k}(x) \big] +
2T_s(x)T_k(x) - T_{s-k}(x).	\label{eq:rodd}
\end{align}
Similarly, if $r \ge 2$ is an even number,
summing up \eqref{eq:rj} for $j=r, r-2, \ldots, 2$ leads to
\begin{align}
	T_{rs+k}(x) = 2\sum_{j=2, \text{even}}^r \big[ T_{js}(x)T_k(x) - T_{(j-1)s}(x)T_{s-k}(x) \big] +
	T_k(x).	\label{eq:reven}
\end{align}
Combining Eqs \eqref{eq:rodd} and \eqref{eq:reven} together, we obtain equation \eqref{eq:cheb_exp}.}
\end{proof}

\begin{proof}[Proof of Lemma \ref{lem:cheb_compon}.]
{First, we rewrite $P^{(k+1)}(x)$ as follows
\begin{align*}
P^{(k+1)}(x)
=
\sum^{s^{k}-1}\limits_{j=0}b_{j}T_{j}(x)  + \sum^{s-1}\limits_{r=1}b_{rs^{k}}T_{rs^{k}}(x) + f_{1}(x) + f_{2}(x),
\end{align*}
where $f_1(x)$, $f_2(x)$ contain all the terms of form $T_{rs^k+j}(x)$, $r\ge 1$, $j=1,\ldots, s^k-1$.
Using Lemma \ref{lem:cheb_expan}, we can write $f_1(x)$ and $f_2(x)$ as
\begin{align*}
f_{1}(x)
=\ &
\sum^{s-1}\limits_{r=1}\sum^{s^{k}-1}_{j=1} b_{r s^{k}+j} \left[\delta^{+}_{r}T_{s^{k}-j}(x)  + \delta^{-}_{r}T_{j}(x)
\right]\nonumber \\
=\ &
\sum^{s^{k}-1}\limits_{j=1}\left[\sum^{s-1}\limits_{r=1}\left(b_{r s^{k}+j}\delta^{-}_{r} + b_{(r+1) s^{k}-j}\delta^{+}_{r}\right) \right]T_{j}(x),
\end{align*}
\begin{align*}
f_{2}(x)
=\ &
2\sum^{s-1}\limits_{r=1}\sum^{s^{k}-1}_{j=1}b_{r s^{k}+j}\sum^{r}_{i=1}\delta^{-}_{r-i} \left[T_{i s^{k}}(x) T_{j}(x) - T_{(i-1)s^{k}}T_{s^{k}-j}(x) \right] \\
={} &
\sum^{s-2}\limits_{i=1}\left\{2\sum^{s^{k}-1}\limits_{j=1}\left[\sum^{s-1}\limits_{r=i}\left(b_{r s^{k}+j}\delta^{-}_{r-i}
-b_{(r+1) s^{k}-j}\delta^{+}_{r-i}\right) \right]T_{j}(x) \right\} T_{i s^{k}}(x) \\
& {} -2\sum^{s^{k}-1}_{j=1}\left(\sum^{s-1}_{r=1} b_{(r+1) s^{k}-j}\delta^{-}_{r-1} \right)T_{j}(x) + 2\left[\sum^{s^{k}-1}_{j=1}b_{(s-1)s^{k}+j}T_{j}(x)  \right] T_{(s-1)s^k}(x).
\end{align*}
By arranging terms, we get
\begin{align*}
P^{(k+1)}(x)
=\ &
b_{0} +
\sum^{s^{k}-1}\limits_{j=1}\left[b_{j} + \sum^{s-1}\limits_{r=1}\Big(b_{r s^{k}+j}\delta^{-}_{r} - b_{(r+1) s^{k}-j}\delta^{+}_{r}\Big) \right]T_{j}(x) \\
\quad  &
+\sum^{s-2}\limits_{i=1}\left\{
b_{i s^{k}} + 2\sum^{s^{k}-1}\limits_{j=1}\left[\sum^{s-1}\limits_{r=i}\left(b_{r s^{k}+j}\delta^{-}_{r-i}-b_{(r+1) s^{k}-j}\delta^{+}_{r-i}\right) \right]T_{j}(x)
\right\} T_{i s^{k}}(x) \\
\quad &
+ \left( b_{(s-1) s^{k}}
+ 2\sum^{s^{k}-1}\limits_{j=1}b_{(s-1) s^{k}+j}
T_{j}(x) \right) T_{(s-1)s^{k}}(x).
\end{align*}
We obtain Eq. \eqref{eq:poly_recur_s2} by using the fact $T_{mn}(x)=T_m(T_n(x))$.}
\end{proof}

\begin{proof}{[sketch of the proof for Theorem \ref{thm:1d-s}]
	According to Corollary 1.1 and Corollary 2.1 in \cite{li_PowerNet_2019},
	polynomials of degree $\le s$ can be accurately represented by  $\sigma_{s}$-neural networks with only one hidden layer, and moreover, polynomials of degree $<s$ with variable coefficients can also be realized by $\sigma_{s}$-neural networks with only one hidden layer. More precisely,
	\begin{align*}
	\sum^{s}\limits_{j=0}d_{j}x^{j} =& {}
	d_0 + \sum_{j=1}^s d_j \left(\gamma^{T}_{1,j}\sigma_{s}(\alpha_{1}x+\beta_{1}) + \lambda_{0,j} \right), \\
	\sum_{t=0}^{s-1} x^{t}y_t = & \sum_{t=0}^{s-1} \gamma^{T}_{2,t}\sigma_{s}(\alpha_{2,t,1}\, x+\alpha_{2,t,2}\, y_t + \beta_{2,t}),
	\end{align*}
	where $\alpha_{1}$, $\beta_{1}$, $\gamma_{1,j}$, $\lambda_{0,j}$,
	$\alpha_{2,t,1}$, $\alpha_{2,t,2}$,  $\beta_{2,t}$,  $\gamma_{2,t}$ are
	constant vectors, {$y_t$ are variable coefficients}.
	Using the above results, we have that
	$\sum^{s}_{j=0}b_{j}T_j(x)$, $\sum_{t=0}^{s-1} y_t T_t(x)  $ can be realized by $\sigma_s$ networks of only one hidden layer having same network structures as in the PowerNet case.

  Eq \eqref{eq:poly_recur_s2} in Lemma \ref{lem:cheb_compon} can be regarded as a polynomials of degree $s-1$ of variable
$T_{s^k}(x)$ after $\widetilde{P}_{l}^{(k)}(x)$, $l=0,1,\ldots, s-1$ and $T_{s^k}(x)$ are calculated out. Thus it can be realized by a $\sigma_s$ network of one hidden layer. Since $T_{s^k}(x)=T_s(T_{s^{k-1}}(x))$, and $\widetilde{P}_{l}^{(k)}(x)$ are polynomials of degree less than $s^k$, the overall calculation of $P^{(k+1)}(x)$ can be realized recursively. The overall network structure is similar to the PowerNet case (see Theorem 2 in \cite{li_PowerNet_2019}). So, for a polynomial of degree at most $n$, the number of total hidden layers in such a recursive realization is $\lceil\log_s n\rceil+1$.
The number of nodes and nonzero weights are $\mathcal{O}(n)$ and $\mathcal{O}(sn)$, respectively. }
\end{proof}

 Similar to $\sigma_2$ case, the recursive realization is equivalent to first transform the representation from Chebyshev polynomial expansion
to hierarchical Chebyshev expansion based on $s$-section, then use the procedure of building $\sigma_s$ PowerNets to build the overall $\sigma_s$ network realization of the Chebyshev expansions. Now we describe how to generate the transform matrix from
standard Chebyshev expansion to hierarchical Chebyshev expansion.

For convenience, let's denote that
\begin{align*}
A^{(k)}_{l,0}
=\
e^{(l s^{k})}_{s,k}, \qquad
A^{(k)}_{l,j}
=\
(2-\delta_{l0})\cdot \sum^{s-1}\limits_{r=l}\Big(\delta^{-}_{r-l}e^{(r s^{k}+j)}_{s,k} - \delta^{+}_{r-l}e^{((r+1) s^{k}-j)}_{s,k}\Big).
\end{align*}
\begin{itemize}
\item For $k=0$, let $P^{(1)}(x)\!=\!\sum^{s-1}_{j=0}b_{j}T_{j}(x)$, Eq \eqref{eq:poly_recur_s2} is
\begin{align*}
P^{(1)}(x)\!=\!B_{0} + B_{1} T_{1}\Big(T_{s^{0}}(x)\Big) + B_{2} T_{2}\Big(T_{s^{0}}(x)\Big) + \cdots + B_{s-1} T_{s-1}\Big(T_{s^{0}}(x)\Big),
\end{align*}
and then
\begin{small}
\begin{align*}
\begin{bmatrix}
B_{0} \\
\vdots \\
B_{s-1}
\end{bmatrix}
\!=
S_{0}
\bm{b}_{0} , \quad
S_{0} =
I_{s}.
\end{align*}
\end{small}
\item For $k=1,\ P^{(2)}(x)\!=\!\sum^{s^{2}-1}_{j=0}b_{j}T_{j}(x)$, according to Lemma \ref{lem:cheb_compon}, we can rewrite it as
\begin{align*}
P^{(2)}(x)
\!=\!
P^{(1)}_{0}(x) + P^{(1)}_{1}(x) T_{1}\Big(T_{s}(x)\Big) + \cdots + P^{(1)}_{s-1}(x) T_{s-1}\Big(T_{s}(x) \Big).
\end{align*}
Since $\deg\big(P^{(1)}_{i_{1}}\big)\!=\!s-1$, so we can write $P_{i_1}^{(1)}$ as
\begin{align*}
P^{(1)}_{i_{1}}(x) = \widetilde{b}^{(i_{1})}_{1,0} \:+\: \widetilde{b}^{(i_{1})}_{1,1} T_{1}(x) \:+\: \widetilde{b}^{(i_{1})}_{1,2} T_{2}(x) \:+\: \cdots \:+\: \widetilde{b}^{(i_{1})}_{1,s-1} T_{s-1}(x),
\end{align*}
where $i_{1}=0,1,\ldots,s-1$. On the other hand,  we have
\begin{align*}
P^{(1)}_{i_{1}}(x) =
B_{0}^{(i_{1})}  + B_{1}^{(i_{1})}  T_{1}\Big(T_{s^{0}}(x)\Big) + \cdots +
B_{s-1}^{(i_{1})}  T_{s-1}\Big(T_{s^{0}}(x) \Big).
\end{align*}
From the conclusion of $k = 0$, and the conclusion in Lemma \ref{lem:cheb_compon}, we get
\begin{small}
\begin{align*}
M^{(i_{1})}
:=
\begin{bmatrix}
B_{0}^{(i_{1})}  \\
\vdots \\
B_{s-1}^{(i_{1})}
\end{bmatrix}
=
S_{0}
\begin{bmatrix}
\widetilde{b}^{(i_{1})}_{1,0} \\
\vdots \\
\widetilde{b}^{(i_{1})}_{1,s-1}
\end{bmatrix},\ \text{and}\
\begin{bmatrix}
\widetilde{b}^{(i_{1})}_{1,0} \\
\vdots \\
\widetilde{b}^{(i_{1})}_{1,s-1}
\end{bmatrix}
=
\begin{bmatrix}
A^{(1)}_{i_{1},0} \\
\vdots \\
A^{(1)}_{i_{1},s-1}
\end{bmatrix}
 \bm{b}_{1}
=: R^{(1)}_{i_{1}} \bm{b}_{1},
\end{align*}
\end{small}
then
\begin{small}
\begin{align*}
\begin{bmatrix}
M^{(0)}  \\
\vdots \\
M^{(s-1)}
\end{bmatrix}
=
\Big(I_{s}\otimes S_{0}\Big)
\begin{bmatrix}
R^{(1)}_{0}  \\
\vdots \\
R^{(1)}_{s-1}
\end{bmatrix}
\bm{b}_{1}
=: S_{1} \bm{b}_{1}.
\end{align*}
\end{small}
\item For $k=2,\ P^{(3)}(x)\!=\!\sum^{s^{3}-1}_{j=0}b_{j}T_{j}(x)$, according to Lemma \ref{lem:cheb_compon}, we can rewrite it as
\begin{align*}
P^{(3)}(x)
\:=\:
P^{(2)}_{0}(x) +
P^{(2)}_{1}(x) T_{1}\Big(T_{s^{2}}(x)\Big) + \cdots +
P^{(2)}_{s-1}(x) T_{s-1}\Big(T_{s^{2}}(x)\Big).
\end{align*}
The condition $\deg\big(P^{(2)}_{i_{2}} \big)=s^{2}-1$ is satisfied at this time, so we can write
\begin{align*}
P^{(2)}_{i_{1}}(x) = \widetilde{b}^{(i_{1})}_{2,0} \:+\: \widetilde{b}^{(i_{1})}_{2,1}T_{1}(x) \:+\: \cdots \:+\: \widetilde{b}^{(i_{1})}_{2,s^{2}-1}T_{s^{2}-1}(x),
\end{align*}
where $i_{1}=0,1,..,s-1$. And on the other hand, by Lemma \ref{lem:cheb_compon} we have,
\begin{align*}
P^{(2)}_{i_{1}}(x) =
\widetilde{P}^{(2)}_{i_{1},0}(x)
+ \widetilde{P}^{(2)}_{i_{1},1}(x)  T_{1}\Big(T_{s}(x)\Big) + \cdots +
\widetilde{P}^{(2)}_{i_{1},s-1}(x)  T_{s-1}\Big(T_{s}(x) \Big),
\end{align*}
where ${\operatorname{deg}}\big(\widetilde{P}^{(2)}_{i_{1},i_{2}}(x) \big)=s-1$,
$i_{2}=0,1,\ldots,s-1$. So, we can write
\begin{align*}
\widetilde{P}^{(2)}_{i_{1},i_{2}}(x)
=\ &
\widetilde{b}^{(i_{1},i_{2})}_{2,0} \:+\: \widetilde{b}^{(i_{1},i_{2})}_{2,1}T_{1 }(x) \:+\: \cdots \:+\: \widetilde{b}^{(i_{1},i_{2})}_{2,s-1}T_{s-1}(x) \\
=\ &
B^{(i_{1},i_{2})}_{0} + B^{(i_{1},i_{2})}_{1}T_{1}\Big(T_{s^{0}}(x)\Big) + \cdots + B^{(i_{1},i_{2})}_{s-1}T_{s-1}\Big(T_{s^{0}}(x)\Big).
\end{align*}
So, from the conclusion of $k = 0$, and  Lemma \ref{lem:cheb_compon}, we obtain
\begin{small}
\begin{align*}
M^{(i_{1},i_{2})}
:=
\begin{bmatrix}
B_{0}^{(i_{1},i_{2})}  \\
\vdots \\
B_{s-1}^{(i_{1},i_{2})}
\end{bmatrix} =
S_{0}
\begin{bmatrix}
\widetilde{b}^{(i_{1},i_{2})}_{2,0} \\
\vdots \\
\widetilde{b}^{(i_{1},i_{2})}_{2,s-1}
\end{bmatrix} =
S_{0} R^{(1)}_{i_{2}}
\begin{bmatrix}
\widetilde{b}^{(i_{1})}_{2,0} \\
\vdots \\
\widetilde{b}^{(i_{1})}_{2,s^{2}-1}
\end{bmatrix}
=
S_{0}R^{(1)}_{i_{2}}R^{(2)}_{i_{1}} \bm{b}_{2},
\end{align*}
\end{small}
and then
\begin{small}
\begin{align*}
\begin{bmatrix}
M^{(0,0)}  \\
\vdots \\
M^{(s-1,s-1)}
\end{bmatrix}
=
\Big(I_{s}\otimes S_{1}\Big) R^{(2)}
\bm{b}_{2}
=:
S_{2} \bm{b}_{2}.
\end{align*}
\end{small}
\item Similarly, for $k \ge 3$, take $P^{(k+1)}(x)=\sum^{s^{k+1}-1}_{j=0}b_{j}T_{j}(x)$, it can be rewritten as
\begin{align*}
P^{(k+1)}(x)
=
P^{(k)}_{0}(x) +
P^{(k)}_{1}(x) T_{1}\Big(T_{s^{k}}(x)\Big) + \cdots +
P^{(k)}_{s-1}(x) T_{s-1}\Big(T_{s^{k}}(x)\Big).
\end{align*}
We get
\begin{align*}
M_{k} =
\begin{bmatrix}
M^{(0,\ldots,0,0)}  \\
M^{(0,\ldots,0,1)}  \\
\vdots \\
M^{(s-1,\ldots,s-1)}
\end{bmatrix}
=
\Big(I_{s}\otimes S_{k-1}\Big) R^{(k)}
\bm{b}_{k}
=:
S_{k} \bm{b}_{k},
\end{align*}
\begin{align*}
M^{(i_{1},i_{2},\ldots,i_{k})} =
\begin{bmatrix}
B^{(i_{1},i_{2},\ldots,i_{k})}_{0} \\
\vdots \\
B^{(i_{1},i_{2},\ldots,i_{k})}_{s-1}
\end{bmatrix}, \quad
R^{(k)} =
\begin{bmatrix}
R^{(k)}_{0} \\
\vdots \\
R^{(k)}_{s-1}
\end{bmatrix}, \quad
R^{(k)}_{i} =
\begin{bmatrix}
A^{(k)}_{i,0} \\
\vdots \\
A^{(k)}_{i,s^{k}-1} \\
\end{bmatrix},
\end{align*}
\end{itemize}
where $i=0,1,\ldots,s-1$.

\begin{remark}
In particular, when $s={2}$, the transform matrix $S_k$ derived above
is equivalent to \eqref{eq:S_recur}.
When $s = 3$, the matrix $S_{k}$ can be expressed as follows
\begin{align}
S_{0} = I_{3}, \qquad S_{k} :=
\Big(I_{3}\otimes S_{k-1}\Big)
\begin{bmatrix}
I_{3^{k}+1} &\ A^{(1)}_{k} &\ A^{(2)}_{k} \\
\bm{0} &\ 2I_{3^{k}}  &\  A^{(3)}_{k} \\
\bm{0} &\ \bm{0} &\ 2I_{3^{k}-1}
\end{bmatrix}, \quad \text{for}\ k\ge 1,
\end{align}
\begin{align}
A^{(1)}_{k} :=
\begin{bmatrix}
-J_{3^{k}} \\
\bm{0}_{1\times 3^{k}}
\end{bmatrix},\quad
A^{(2)}_{k} :=
\begin{bmatrix}
\bm{0}_{1\times (3^{k}-1)} \\
I_{3^{k}-1}  \\
\bm{0}_{1\times (3^{k}-1)}
\end{bmatrix},\quad
A^{(3)}_{k} :=
\begin{bmatrix}
\bm{0}_{1\times (3^{k}-1)} \\
-2J_{3^{k}-1}
\end{bmatrix},
\end{align}
where $I_{n}, J_{n}$ denotes the unit matrix of order $n$ and the oblique diagonal unit matrix of order $n$, respectively.
\end{remark}

\end{appendices}

\bibliography{ChebNet}


\begin{thebibliography}{56}
\ifx \bisbn   \undefined \def \bisbn  #1{ISBN #1}\fi
\ifx \binits  \undefined \def \binits#1{#1}\fi
\ifx \bauthor  \undefined \def \bauthor#1{#1}\fi
\ifx \batitle  \undefined \def \batitle#1{#1}\fi
\ifx \bjtitle  \undefined \def \bjtitle#1{#1}\fi
\ifx \bvolume  \undefined \def \bvolume#1{\textbf{#1}}\fi
\ifx \byear  \undefined \def \byear#1{#1}\fi
\ifx \bissue  \undefined \def \bissue#1{#1}\fi
\ifx \bfpage  \undefined \def \bfpage#1{#1}\fi
\ifx \blpage  \undefined \def \blpage #1{#1}\fi
\ifx \burl  \undefined \def \burl#1{\textsf{#1}}\fi
\ifx \doiurl  \undefined \def \doiurl#1{\url{https://doi.org/#1}}\fi
\ifx \betal  \undefined \def \betal{\textit{et al.}}\fi
\ifx \binstitute  \undefined \def \binstitute#1{#1}\fi
\ifx \binstitutionaled  \undefined \def \binstitutionaled#1{#1}\fi
\ifx \bctitle  \undefined \def \bctitle#1{#1}\fi
\ifx \beditor  \undefined \def \beditor#1{#1}\fi
\ifx \bpublisher  \undefined \def \bpublisher#1{#1}\fi
\ifx \bbtitle  \undefined \def \bbtitle#1{#1}\fi
\ifx \bedition  \undefined \def \bedition#1{#1}\fi
\ifx \bseriesno  \undefined \def \bseriesno#1{#1}\fi
\ifx \blocation  \undefined \def \blocation#1{#1}\fi
\ifx \bsertitle  \undefined \def \bsertitle#1{#1}\fi
\ifx \bsnm \undefined \def \bsnm#1{#1}\fi
\ifx \bsuffix \undefined \def \bsuffix#1{#1}\fi
\ifx \bparticle \undefined \def \bparticle#1{#1}\fi
\ifx \barticle \undefined \def \barticle#1{#1}\fi
\bibcommenthead
\ifx \bconfdate \undefined \def \bconfdate #1{#1}\fi
\ifx \botherref \undefined \def \botherref #1{#1}\fi
\ifx \url \undefined \def \url#1{\textsf{#1}}\fi
\ifx \bchapter \undefined \def \bchapter#1{#1}\fi
\ifx \bbook \undefined \def \bbook#1{#1}\fi
\ifx \bcomment \undefined \def \bcomment#1{#1}\fi
\ifx \oauthor \undefined \def \oauthor#1{#1}\fi
\ifx \citeauthoryear \undefined \def \citeauthoryear#1{#1}\fi
\ifx \endbibitem  \undefined \def \endbibitem {}\fi
\ifx \bconflocation  \undefined \def \bconflocation#1{#1}\fi
\ifx \arxivurl  \undefined \def \arxivurl#1{\textsf{#1}}\fi
\csname PreBibitemsHook\endcsname

\bibitem{hinton_fast_2006}
\begin{barticle}
\bauthor{\bsnm{Hinton}, \binits{G.}},
\bauthor{\bsnm{Osindero}, \binits{S.}},
\bauthor{\bsnm{Teh}, \binits{Y.-W.}}:
\batitle{A fast learning algorithm for deep belief nets}.
\bjtitle{Neural Computation}
\bvolume{18}(\bissue{7}),
\bfpage{1527}--\blpage{1554}
(\byear{2006}).
\doiurl{10.1162/neco.2006.18.7.1527}
\end{barticle}
\endbibitem

\bibitem{bengio_greedy_2007}
\begin{bchapter}
\bauthor{\bsnm{Bengio}, \binits{Y.}},
\bauthor{\bsnm{Lamblin}, \binits{P.}},
\bauthor{\bsnm{Popovici}, \binits{D.}},
\bauthor{\bsnm{Larochelle}, \binits{H.}}:
\bctitle{Greedy layer-wise training of deep networks}.
In: \bbtitle{Advances in Neural Information Processing Systems},
pp. \bfpage{153}--\blpage{160}
(\byear{2007})
\end{bchapter}
\endbibitem

\bibitem{hinton_deep_2012}
\begin{botherref}
\oauthor{\bsnm{Hinton}, \binits{G.}},
\oauthor{\bsnm{Deng}, \binits{L.}},
\oauthor{\bsnm{Yu}, \binits{D.}},
\oauthor{\bsnm{Dahl}, \binits{G.}},
\oauthor{\bsnm{Mohamed}, \binits{A.}},
\oauthor{\bsnm{Jaitly}, \binits{N.}},
\oauthor{\bsnm{Senior}, \binits{A.}},
\oauthor{\bsnm{Vanhoucke}, \binits{V.}},
\oauthor{\bsnm{Nguyen}, \binits{P.}},
\oauthor{\bsnm{Kingsbury}, \binits{B.}},
\oauthor{\bsnm{Sainath}, \binits{T.}}:
Deep neural networks for acoustic modeling in speech recognition.
IEEE Signal Process. Mag.
\textbf{29}
(2012)
\end{botherref}
\endbibitem

\bibitem{krizhevsky_imagenet_2012}
\begin{barticle}
\bauthor{\bsnm{Krizhevsky}, \binits{A.}},
\bauthor{\bsnm{Sutskever}, \binits{I.}},
\bauthor{\bsnm{Hinton}, \binits{G.E.}}:
\batitle{{ImageNet} classification with deep convolutional neural networks}.
\bjtitle{Neural Information Processing Systems}
\bvolume{141}(\bissue{5}),
\bfpage{1097}--\blpage{1105}
(\byear{2012})
\end{barticle}
\endbibitem

\bibitem{lecun_deep_2015}
\begin{barticle}
\bauthor{\bsnm{LeCun}, \binits{Y.}},
\bauthor{\bsnm{Bengio}, \binits{Y.}},
\bauthor{\bsnm{Hinton}, \binits{G.}}:
\batitle{Deep learning}.
\bjtitle{Nature}
\bvolume{521}(\bissue{7553}),
\bfpage{436}--\blpage{444}
(\byear{2015})
\end{barticle}
\endbibitem

\bibitem{he2016deep}
\begin{bchapter}
\bauthor{\bsnm{He}, \binits{K.}},
\bauthor{\bsnm{Zhang}, \binits{X.}},
\bauthor{\bsnm{Ren}, \binits{S.}},
\bauthor{\bsnm{Sun}, \binits{J.}}:
\bctitle{Deep residual learning for image recognition}.
In: \bbtitle{Proceedings of the IEEE Conference on Computer Vision and Pattern
  Recognition},
pp. \bfpage{770}--\blpage{778}
(\byear{2016})
\end{bchapter}
\endbibitem

\bibitem{zhang_deep_2018}
\begin{barticle}
\bauthor{\bsnm{Zhang}, \binits{L.}},
\bauthor{\bsnm{Han}, \binits{J.}},
\bauthor{\bsnm{Wang}, \binits{H.}},
\bauthor{\bsnm{Car}, \binits{R.}},
\bauthor{\bsnm{E}, \binits{W.}}:
\batitle{Deep potential molecular dynamics: A scalable model with the accuracy
  of quantum mechanics}.
\bjtitle{Phys. Rev. Lett.}
\bvolume{120}(\bissue{14}),
\bfpage{143001}
(\byear{2018})
\end{barticle}
\endbibitem

\bibitem{e_deep_2018}
\begin{barticle}
\bauthor{\bsnm{E}, \binits{W.}},
\bauthor{\bsnm{Yu}, \binits{B.}}:
\batitle{The deep {Ritz} method: A deep learning-based numerical algorithm for
  solving variational problems}.
\bjtitle{Commun. Math. Stat.}
\bvolume{6}(\bissue{1}),
\bfpage{1}--\blpage{12}
(\byear{2018}).
\doiurl{10.1007/s40304-018-0127-z}
\end{barticle}
\endbibitem

\bibitem{han_solving_2018}
\begin{barticle}
\bauthor{\bsnm{Han}, \binits{J.}},
\bauthor{\bsnm{Jentzen}, \binits{A.}},
\bauthor{\bsnm{E}, \binits{W.}}:
\batitle{Solving high-dimensional partial differential equations using deep
  learning}.
\bjtitle{Proceedings of the National Academy of Sciences}
\bvolume{115}(\bissue{34}),
\bfpage{8505}--\blpage{8510}
(\byear{2018}).
\doiurl{10.1073/pnas.1718942115}
\end{barticle}
\endbibitem

\bibitem{kutyniok_theoretical_2019}
\begin{barticle}
\bauthor{\bsnm{Kutyniok}, \binits{G.}},
\bauthor{\bsnm{Petersen}, \binits{P.}},
\bauthor{\bsnm{Raslan}, \binits{M.}},
\bauthor{\bsnm{Schneider}, \binits{R.}}:
\batitle{A theoretical analysis of deep neural networks and parametric {PDEs}}.
\bjtitle{Constructive Approximation}
\bvolume{55},
\bfpage{73}--\blpage{125}
(\byear{2022})
{\href{https://arxiv.org/abs/1904.00377}{{arXiv:1904.00377}}}
\end{barticle}
\endbibitem

\bibitem{cybenko_approximation_1989}
\begin{barticle}
\bauthor{\bsnm{Cybenko}, \binits{G.}}:
\batitle{Approximation by superpositions of a sigmoidal function}.
\bjtitle{Math. Control Signal Systems}
\bvolume{2}(\bissue{4}),
\bfpage{303}--\blpage{314}
(\byear{1989}).
\doiurl{10.1007/BF02551274}
\end{barticle}
\endbibitem

\bibitem{hornik_multilayer_1989}
\begin{barticle}
\bauthor{\bsnm{Hornik}, \binits{K.}},
\bauthor{\bsnm{Stinchcombe}, \binits{M.}},
\bauthor{\bsnm{White}, \binits{H.}}:
\batitle{Multilayer feedforward networks are universal approximators}.
\bjtitle{Neural Networks}
\bvolume{2}(\bissue{5}),
\bfpage{359}--\blpage{366}
(\byear{1989})
\end{barticle}
\endbibitem

\bibitem{mhaskar_neural_1996}
\begin{barticle}
\bauthor{\bsnm{Mhaskar}, \binits{H.N.}}:
\batitle{Neural networks for optimal approximation of smooth and analytic
  functions}.
\bjtitle{Neural Computation}
\bvolume{8}(\bissue{1}),
\bfpage{164}--\blpage{177}
(\byear{1996}).
\doiurl{10.1162/neco.1996.8.1.164}
\end{barticle}
\endbibitem

\bibitem{telgarsky_representation_2015}
\begin{botherref}
\oauthor{\bsnm{Telgarsky}, \binits{M.}}:
Representation benefits of deep feedforward networks.
ArXiv150908101 Cs
(2015)
\end{botherref}
\endbibitem

\bibitem{eldan_power_2016}
\begin{barticle}
\bauthor{\bsnm{Eldan}, \binits{R.}},
\bauthor{\bsnm{Shamir}, \binits{O.}}:
\batitle{The power of depth for feedforward neural networks}.
\bjtitle{JMLR Workshop Conf. Proc.}
\bvolume{49},
\bfpage{1}--\blpage{34}
(\byear{2016})
\end{barticle}
\endbibitem

\bibitem{poggio_why_2017}
\begin{barticle}
\bauthor{\bsnm{Poggio}, \binits{T.}},
\bauthor{\bsnm{Mhaskar}, \binits{H.}},
\bauthor{\bsnm{Rosasco}, \binits{L.}},
\bauthor{\bsnm{Miranda}, \binits{B.}},
\bauthor{\bsnm{Liao}, \binits{Q.}}:
\batitle{Why and when can deep-but not shallow-networks avoid the curse of
  dimensionality: {{A}} review}.
\bjtitle{Int. J. Autom. Comput.}
\bvolume{14}(\bissue{5}),
\bfpage{503}--\blpage{519}
(\byear{2017}).
\doiurl{10.1007/s11633-017-1054-2}
\end{barticle}
\endbibitem

\bibitem{liang_why_2016}
\begin{botherref}
\oauthor{\bsnm{Liang}, \binits{S.}},
\oauthor{\bsnm{Srikant}, \binits{R.}}:
Why deep neural networks for function approximation?
arXiv:1610.04161, ICLR
(2017)
{\href{https://arxiv.org/abs/1610.04161}{{arXiv:1610.04161}}}
{[cs]}
\end{botherref}
\endbibitem

\bibitem{telgarsky_benefits_2016a}
\begin{bchapter}
\bauthor{\bsnm{Telgarsky}, \binits{M.}}:
\bctitle{Benefits of depth in neural networks}.
In: \bbtitle{JMLR: Workshop and Conference Proceedings},
vol. \bseriesno{49},
pp. \bfpage{1}--\blpage{23}
(\byear{2016})
\end{bchapter}
\endbibitem

\bibitem{yarotsky_error_2017}
\begin{barticle}
\bauthor{\bsnm{Yarotsky}, \binits{D.}}:
\batitle{Error bounds for approximations with deep {ReLU} networks}.
\bjtitle{Neural Networks}
\bvolume{94},
\bfpage{103}--\blpage{114}
(\byear{2017}).
\doiurl{10.1016/j.neunet.2017.07.002}
\end{barticle}
\endbibitem

\bibitem{petersen_optimal_2018}
\begin{barticle}
\bauthor{\bsnm{Petersen}, \binits{P.}},
\bauthor{\bsnm{Voigtlaender}, \binits{F.}}:
\batitle{Optimal approximation of piecewise smooth functions using deep
  {{ReLU}} neural networks}.
\bjtitle{Neural Networks}
\bvolume{108},
\bfpage{296}--\blpage{330}
(\byear{2018}).
\doiurl{10.1016/j.neunet.2018.08.019}
\end{barticle}
\endbibitem

\bibitem{e_exponential_2018}
\begin{barticle}
\bauthor{\bsnm{E}, \binits{W.}},
\bauthor{\bsnm{Wang}, \binits{Q.}}:
\batitle{Exponential convergence of the deep neural network approximation for
  analytic functions}.
\bjtitle{Sci. China Math.}
\bvolume{61}(\bissue{10}),
\bfpage{1733}--\blpage{1740}
(\byear{2018})
\end{barticle}
\endbibitem

\bibitem{opschoor_exponential_2019}
\begin{barticle}
\bauthor{\bsnm{Opschoor}, \binits{J.A.A.}},
\bauthor{\bsnm{Schwab}, \binits{C.}},
\bauthor{\bsnm{Zech}, \binits{J.}}:
\batitle{Exponential {ReLU} {DNN} expression of holomorphic maps in high
  dimension}.
\bjtitle{Constructive Approximation}
(\byear{2021}).
\doiurl{10.1007/s00365-021-09542-5}
\end{barticle}
\endbibitem

\bibitem{li_better_2019}
\begin{barticle}
\bauthor{\bsnm{Li}, \binits{B.}},
\bauthor{\bsnm{Tang}, \binits{S.}},
\bauthor{\bsnm{Yu}, \binits{H.}}:
\batitle{Better approximations of high dimensional smooth functions by deep
  neural networks with rectified power units}.
\bjtitle{Commun. Comput. Phys.}
\bvolume{27}(\bissue{2}),
\bfpage{379}--\blpage{411}
(\byear{2020})
\end{barticle}
\endbibitem

\bibitem{mhaskar_approximation_1993}
\begin{barticle}
\bauthor{\bsnm{Mhaskar}, \binits{H.N.}}:
\batitle{Approximation properties of a multilayered feedforward artificial
  neural network}.
\bjtitle{Adv. Comput. Math.}
\bvolume{1}(\bissue{1}),
\bfpage{61}--\blpage{80}
(\byear{1993})
\end{barticle}
\endbibitem

\bibitem{li_PowerNet_2019}
\begin{barticle}
\bauthor{\bsnm{Li}, \binits{B.}},
\bauthor{\bsnm{Tang}, \binits{S.}},
\bauthor{\bsnm{Yu}, \binits{H.}}:
\batitle{{PowerNet}: Efficient representations of polynomials and smooth
  functions by deep neural networks with rectified power units}.
\bjtitle{arXiv:1909.05136, J. Math. Study}
\bvolume{53}(\bissue{2}),
\bfpage{159}--\blpage{191}
(\byear{2020})
\end{barticle}
\endbibitem

\bibitem{raissi_physicsinformed_2019}
\begin{barticle}
\bauthor{\bsnm{Raissi}, \binits{M.}},
\bauthor{\bsnm{Perdikaris}, \binits{P.}},
\bauthor{\bsnm{Karniadakis}, \binits{G.E.}}:
\batitle{Physics-informed neural networks: {{A}} deep learning framework for
  solving forward and inverse problems involving nonlinear partial differential
  equations}.
\bjtitle{J. Comput. Phys.}
\bvolume{378},
\bfpage{686}--\blpage{707}
(\byear{2019}).
\doiurl{10.1016/j.jcp.2018.10.045}
\end{barticle}
\endbibitem

\bibitem{xu_finite_2020a}
\begin{barticle}
\bauthor{\bsnm{Xu}, \binits{J.}}:
\batitle{Finite neuron method and convergence analysis}.
\bjtitle{Commun. Comput. Phys.}
\bvolume{28}(\bissue{5}),
\bfpage{1707}--\blpage{1745}
(\byear{2020}).
\doiurl{10.4208/cicp.OA-2020-0191}
\end{barticle}
\endbibitem

\bibitem{yu_onsagernet_2021}
\begin{barticle}
\bauthor{\bsnm{Yu}, \binits{H.}},
\bauthor{\bsnm{Tian}, \binits{X.}},
\bauthor{\bsnm{E}, \binits{W.}},
\bauthor{\bsnm{Li}, \binits{Q.}}:
\batitle{{{OnsagerNet}}: {{Learning}} stable and interpretable dynamics using a
  generalized {{Onsager}} principle}.
\bjtitle{Phys. Rev. Fluids}
\bvolume{6}(\bissue{11}),
\bfpage{114402}
(\byear{2021})
{\href{https://arxiv.org/abs/2009.02327}{{arxiv:2009.02327}}}.
\doiurl{10.1103/PhysRevFluids.6.114402}
\end{barticle}
\endbibitem

\bibitem{lyu_mim_2022}
\begin{barticle}
\bauthor{\bsnm{Lyu}, \binits{L.}},
\bauthor{\bsnm{Zhang}, \binits{Z.}},
\bauthor{\bsnm{Chen}, \binits{M.}},
\bauthor{\bsnm{Chen}, \binits{J.}}:
\batitle{{{MIM}}: {{A}} deep mixed residual method for solving high-order
  partial differential equations}.
\bjtitle{J. Comput. Phys.}
\bvolume{452},
\bfpage{110930}
(\byear{2022}).
\doiurl{10.1016/j.jcp.2021.110930}
\end{barticle}
\endbibitem

\bibitem{gautschi_optimally_2011}
\begin{barticle}
\bauthor{\bsnm{Gautschi}, \binits{W.}}:
\batitle{Optimally scaled and optimally conditioned {V}andermonde and
  {V}andermonde-like matrices}.
\bjtitle{{BIT} Numerical Mathematics}
\bvolume{51}(\bissue{1}),
\bfpage{103}--\blpage{125}
(\byear{2011}).
\doiurl{10.1007/s10543-010-0293-1}
\end{barticle}
\endbibitem

\bibitem{higham_error_1987}
\begin{barticle}
\bauthor{\bsnm{Higham}, \binits{N.J.}}:
\batitle{Error analysis of the {{Bj{\"o}rck}}-{{Pereyra}} algorithms for
  solving {{Vandermonde}} systems}.
\bjtitle{Numer. Math.}
\bvolume{50}(\bissue{5}),
\bfpage{613}--\blpage{632}
(\byear{1987}).
\doiurl{10.1007/BF01408579}
\end{barticle}
\endbibitem

\bibitem{boyd_chebyshev_2001}
\begin{bbook}
\bauthor{\bsnm{Boyd}, \binits{J.P.}}:
\bbtitle{{C}hebyshev and {F}ourier Spectral Methods},
\bedition{2}nd edn.
\bpublisher{Dover Publications, INC.},
\blocation{New York}
(\byear{2000})
\end{bbook}
\endbibitem

\bibitem{platte_chebfun_2010}
\begin{bchapter}
\bauthor{\bsnm{Platte}, \binits{R.B.}},
\bauthor{\bsnm{Trefethen}, \binits{L.N.}}:
\bctitle{Chebfun: {{A New Kind}} of {{Numerical Computing}}}.
In: \beditor{\bsnm{Fitt}, \binits{A.D.}},
\beditor{\bsnm{Norbury}, \binits{J.}},
\beditor{\bsnm{Ockendon}, \binits{H.}},
\beditor{\bsnm{Wilson}, \binits{E.}} (eds.)
\bbtitle{Progress in Industrial Mathematics at {ECMI} 2008}.
\bsertitle{Mathematics in {{Industry}}},
pp. \bfpage{69}--\blpage{87}.
\bpublisher{{Springer}},
\blocation{{Berlin, Heidelberg}}
(\byear{2010}).
\doiurl{10.1007/978-3-642-12110-4_5}
\end{bchapter}
\endbibitem

\bibitem{defferrard_convolutional_2016}
\begin{botherref}
\oauthor{\bsnm{Defferrard}, \binits{M.}},
\oauthor{\bsnm{Bresson}, \binits{X.}},
\oauthor{\bsnm{Vandergheynst}, \binits{P.}}:
Convolutional neural networks on graphs with fast localized spectral filtering.
Advances in Neural Information Processing Systems 29,
3844--3852
(2016)
{\href{https://arxiv.org/abs/1606.09375}{{arXiv:1606.09375}}}
\end{botherref}
\endbibitem

\bibitem{wang_why_2005}
\begin{barticle}
\bauthor{\bsnm{Wang}, \binits{X.}},
\bauthor{\bsnm{Sloan}, \binits{I.}}:
\batitle{Why are high-dimensional finance problems often of low effective
  dimension?}
\bjtitle{SIAM J. Sci. Comput.}
\bvolume{27}(\bissue{1}),
\bfpage{159}--\blpage{183}
(\byear{2005}).
\doiurl{10.1137/S1064827503429429}
\end{barticle}
\endbibitem

\bibitem{Yserentant2004}
\begin{barticle}
\bauthor{\bsnm{Yserentant}, \binits{H.}}:
\batitle{On the regularity of the electronic {S}chr\"odinger equation in
  {{Hilbert}} spaces of mixed derivatives}.
\bjtitle{Numer. Math.}
\bvolume{98}(\bissue{4}),
\bfpage{731}--\blpage{759}
(\byear{2004})
\end{barticle}
\endbibitem

\bibitem{smolyak_quadrature_1963}
\begin{barticle}
\bauthor{\bsnm{Smolyak}, \binits{S.A.}}:
\batitle{Quadrature and interpolation formulas for tensor products of certain
  classes of functions}.
\bjtitle{Dokl Akad Nauk SSSR}
\bvolume{148}(\bissue{5}),
\bfpage{1042}--\blpage{1045}
(\byear{1963})
\end{barticle}
\endbibitem

\bibitem{bungartz_sparse_2004}
\begin{barticle}
\bauthor{\bsnm{Bungartz}, \binits{H.J.}},
\bauthor{\bsnm{Griebel}, \binits{M.}}:
\batitle{Sparse grids}.
\bjtitle{Acta Numer.}
\bvolume{13},
\bfpage{1}--\blpage{123}
(\byear{2004})
\end{barticle}
\endbibitem

\bibitem{barthelmann_high_2000}
\begin{barticle}
\bauthor{\bsnm{Barthelmann}, \binits{V.}},
\bauthor{\bsnm{Novak}, \binits{E.}},
\bauthor{\bsnm{Ritter}, \binits{K.}}:
\batitle{High dimensional polynomial interpolation on sparse grids}.
\bjtitle{Adv. Comput. Math.}
\bvolume{12}(\bissue{4}),
\bfpage{273}--\blpage{288}
(\byear{2000})
\end{barticle}
\endbibitem

\bibitem{shen_sparse_2010}
\begin{barticle}
\bauthor{\bsnm{Shen}, \binits{J.}},
\bauthor{\bsnm{Wang}, \binits{L.L.}}:
\batitle{Sparse spectral approximations of high-dimensional problems based on
  hyperbolic cross}.
\bjtitle{SIAM J Numer Anal}
\bvolume{48}(\bissue{4}),
\bfpage{1087}--\blpage{1109}
(\byear{2010})
\end{barticle}
\endbibitem

\bibitem{shen_approximations_2014}
\begin{barticle}
\bauthor{\bsnm{Shen}, \binits{J.}},
\bauthor{\bsnm{Wang}, \binits{L.L.}},
\bauthor{\bsnm{Yu}, \binits{H.}}:
\batitle{Approximations by orthonormal mapped {{Chebyshev}} functions for
  higher-dimensional problems in unbounded domains}.
\bjtitle{J. Comput. Appl. Math.}
\bvolume{265},
\bfpage{264}--\blpage{275}
(\byear{2014})
\end{barticle}
\endbibitem

\bibitem{bungartz_adaptive_1992}
\begin{bchapter}
\bauthor{\bsnm{Bungartz}, \binits{H.J.}}:
\bctitle{An adaptive {Poisson} solver using hierarchical bases and sparse
  grids}.
In: \bbtitle{Iterative Methods in Linear Algebra},
\bconflocation{{Brussels, Belgium}},
pp. \bfpage{293}--\blpage{310}
(\byear{1992})
\end{bchapter}
\endbibitem

\bibitem{lin_sparse_2001}
\begin{barticle}
\bauthor{\bsnm{Lin}, \binits{Q.}},
\bauthor{\bsnm{Yan}, \binits{N.}},
\bauthor{\bsnm{Zhou}, \binits{A.}}:
\batitle{A sparse finite element method with high accuracy: {Part I}}.
\bjtitle{Numer. Math.}
\bvolume{88}(\bissue{4}),
\bfpage{731}--\blpage{742}
(\byear{2001}).
\doiurl{10.1007/PL00005456}
\end{barticle}
\endbibitem

\bibitem{shen_efficient_2010}
\begin{barticle}
\bauthor{\bsnm{Shen}, \binits{J.}},
\bauthor{\bsnm{Yu}, \binits{H.}}:
\batitle{Efficient spectral sparse grid methods and applications to
  high-dimensional elliptic problems}.
\bjtitle{SIAM J. Sci. Comput.}
\bvolume{32}(\bissue{6}),
\bfpage{3228}--\blpage{3250}
(\byear{2010})
\end{barticle}
\endbibitem

\bibitem{shen_efficient_2012}
\begin{barticle}
\bauthor{\bsnm{Shen}, \binits{J.}},
\bauthor{\bsnm{Yu}, \binits{H.}}:
\batitle{Efficient spectral sparse grid methods and applications to
  high-dimensional elliptic equations {II}: {{Unbounded}} domains}.
\bjtitle{SIAM J. Sci. Comput.}
\bvolume{34}(\bissue{2}),
\bfpage{1141}--\blpage{1164}
(\byear{2012})
\end{barticle}
\endbibitem

\bibitem{guo_sparse_2016}
\begin{barticle}
\bauthor{\bsnm{Guo}, \binits{W.}},
\bauthor{\bsnm{Cheng}, \binits{Y.}}:
\batitle{A sparse grid discontinuous galerkin method for high-dimensional
  transport equations and its application to kinetic simulations}.
\bjtitle{SIAM J. Sci. Comput.}
\bvolume{38}(\bissue{6}),
\bfpage{3381}--\blpage{3409}
(\byear{2016})
\end{barticle}
\endbibitem

\bibitem{rong_nodal_2017}
\begin{barticle}
\bauthor{\bsnm{Rong}, \binits{Z.}},
\bauthor{\bsnm{Shen}, \binits{J.}},
\bauthor{\bsnm{Yu}, \binits{H.}}:
\batitle{A nodal sparse grid spectral element method for multi-dimensional
  elliptic partial differential equations}.
\bjtitle{Int. J. Numer. Anal. Model.}
\bvolume{14}(\bissue{4-5}),
\bfpage{762}--\blpage{783}
(\byear{2017})
\end{barticle}
\endbibitem

\bibitem{griebel_sparse_2007}
\begin{barticle}
\bauthor{\bsnm{Griebel}, \binits{M.}},
\bauthor{\bsnm{Hamaekers}, \binits{J.}}:
\batitle{Sparse grids for the {{Schr\"odinger}} equation}.
\bjtitle{Math. Model. Numer. Anal.}
\bvolume{41}(\bissue{2}),
\bfpage{215}--\blpage{247}
(\byear{2007})
\end{barticle}
\endbibitem

\bibitem{avila_solving_2013}
\begin{barticle}
\bauthor{\bsnm{Avila}, \binits{G.}},
\bauthor{\bsnm{Carrington}, \binits{T.}}:
\batitle{Solving the {S}chr\"odinger equation using {Smolyak} interpolants}.
\bjtitle{J. Chem. Phys.}
\bvolume{139}(\bissue{13}),
\bfpage{134114}
(\byear{2013})
\end{barticle}
\endbibitem

\bibitem{shen_efficient_2016}
\begin{bchapter}
\bauthor{\bsnm{Shen}, \binits{J.}},
\bauthor{\bsnm{Wang}, \binits{Y.}},
\bauthor{\bsnm{Yu}, \binits{H.}}:
\bctitle{Efficient spectral-element methods for the electronic
  {{Schr\"odinger}} equation}.
In: \beditor{\bsnm{Garcke}, \binits{J.}},
\beditor{\bsnm{Pfl\"uger}, \binits{D.}} (eds.)
\bbtitle{Sparse {{Grids}} and {{Applications}} - {{Stuttgart}} 2014}.
\bsertitle{Lecture {{Notes}} in {{Computational Science}} and {{Engineering}}},
pp. \bfpage{265}--\blpage{289}.
\bpublisher{{Springer International Publishing}},
\blocation{Stuttgart}
(\byear{2016})
\end{bchapter}
\endbibitem

\bibitem{schwab_sparse_2003a}
\begin{barticle}
\bauthor{\bsnm{Schwab}, \binits{C.}},
\bauthor{\bsnm{Todor}, \binits{R.A.}}:
\batitle{Sparse finite elements for elliptic problems with stochastic loading}.
\bjtitle{Numerische Mathematik}
\bvolume{95}(\bissue{4}),
\bfpage{707}--\blpage{734}
(\byear{2003})
\end{barticle}
\endbibitem

\bibitem{nobile_sparse_2008}
\begin{barticle}
\bauthor{\bsnm{Nobile}, \binits{F.}},
\bauthor{\bsnm{Tempone}, \binits{R.}},
\bauthor{\bsnm{Webster}, \binits{C.}}:
\batitle{A sparse grid stochastic collocation method for partial differential
  equations with random input data}.
\bjtitle{SIAM J. Numer. Anal.}
\bvolume{46}(\bissue{5}),
\bfpage{2309}--\blpage{2345}
(\byear{2008}).
\doiurl{10.1137/060663660}
\end{barticle}
\endbibitem

\bibitem{montanelli_deep_2017}
\begin{barticle}
\bauthor{\bsnm{Montanelli}, \binits{H.}},
\bauthor{\bsnm{Du}, \binits{Q.}}:
\batitle{New error bounds for deep {ReLU} networks using sparse grids}.
\bjtitle{SIAM J. Math. Data Sci.}
\bvolume{1}(\bissue{1}),
\bfpage{78}--\blpage{92}
(\byear{2019})
\end{barticle}
\endbibitem

\bibitem{cohen_approximation_2015}
\begin{barticle}
\bauthor{\bsnm{Cohen}, \binits{A.}},
\bauthor{\bsnm{DeVore}, \binits{R.}}:
\batitle{Approximation of high-dimensional parametric {{PDEs}}}.
\bjtitle{Acta Numer.}
\bvolume{24},
\bfpage{1}--\blpage{159}
(\byear{2015})
\end{barticle}
\endbibitem

\bibitem{trefethen_spectral_2000}
\begin{bbook}
\bauthor{\bsnm{Trefethen}, \binits{L.N.}}:
\bbtitle{Spectral Methods in {{MATLAB}}}.
\bsertitle{Software, {{Environments}}, and {{Tools}}}.
\bpublisher{{Society for Industrial and Applied Mathematics (SIAM)}},
\blocation{{Philadelphia, PA}}
(\byear{2000})
\end{bbook}
\endbibitem

\bibitem{cheney_introduction_1982}
\begin{bbook}
\bauthor{\bsnm{Cheney}, \binits{E.W.}}:
\bbtitle{Introduction to Approximation Theory},
\bedition{2nd} edn.
\bpublisher{{AMS Chelsea Publishing}},
\blocation{Rhode Island}
(\byear{1982})
\end{bbook}
\endbibitem

\end{thebibliography}

\end{document}